\documentclass{article}

\usepackage[final,nonatbib]{neurips_2020}

\usepackage[utf8]{inputenc} % allow utf-8 input
\usepackage[T1]{fontenc}    % use 8-bit T1 fonts
\usepackage{hyperref}       % hyperlinks
\usepackage{url}            % simple URL typesetting
\usepackage{booktabs}       % professional-quality tables
\usepackage{amsfonts}       % blackboard math symbols
\usepackage{nicefrac}       % compact symbols for 1/2, etc.
\usepackage{microtype}      % microtypography

\usepackage{graphicx}
\usepackage{subfigure}
\usepackage[utf8]{inputenc} % allow utf-8 input
\usepackage[T1]{fontenc}    % use 8-bit T1 fonts
\usepackage{url}            % simple URL typesetting
\usepackage{amsmath}
\usepackage{amsfonts}       % blackboard math symbols
\usepackage{nicefrac}       % compact symbols for 1/2, etc.

\usepackage{comment}
\usepackage{pdflscape}
\usepackage{algorithm}
\usepackage{algorithmic}
\usepackage{bm}
\usepackage{soul}
\usepackage{multirow}

\usepackage{subfigure}
\usepackage{pdflscape}
\usepackage{color,soul}
\usepackage[normalem]{ulem}

\usepackage{float}

\usepackage{wrapfig}
\usepackage{framed}
\usepackage{xcolor}

\title{Recursive Inference %Mixture Inference 
for Variational Autoencoders}

\author{%
  Minyoung Kim$^1$%\thanks{Use footnote for providing further information about author (webpage, alternative address)---\emph{not} for acknowledging funding agencies.} 
  \\
  $^1$Samsung AI Center \\
  Cambridge, UK \\
  \texttt{mikim21@gmail.com} \\
  \And
  Vladimir Pavlovic$^{1,2}$ \\
  $^2$Rutgers University\\
  Piscataway, NJ, USA \\
  \texttt{vladimir@cs.rutgers.edu} \\
}

\begin{document}

\maketitle

%\vspace{-2.0em}
\begin{abstract}
%\vspace{-0.5em}
Inference networks of traditional Variational Autoencoders (VAEs) are typically amortized, resulting in relatively inaccurate posterior approximation compared to instance-wise variational optimization. Recent semi-amortized approaches were proposed to address this drawback; however, their iterative gradient update procedures can be computationally demanding. %In this paper, we consider a different approach of building a mixture inference model. 
To address these issues, in this paper we introduce an accurate amortized inference algorithm. 
We propose a novel recursive mixture estimation algorithm for VAEs that iteratively augments the current mixture with new components so as to maximally reduce the divergence between the variational and the true posteriors. Using the functional gradient approach, we devise an intuitive learning criteria for selecting a new mixture component: the new component has to improve the data likelihood (lower bound) and, at the same time, be as divergent from the current mixture distribution as possible, thus increasing representational diversity. 
Compared to recently proposed boosted variational inference (BVI), our method relies on amortized inference %\hl{this may be a sticky point; we say above that one should not use amortazied inference because it is imprecise, but here we say we use it} 
in contrast to BVI's non-amortized single optimization instance. %, inappropriate for VAEs. %Also, the regularization to avoid degenerate solution in finding a new component that maximizes divergence from the current mixture is adapted accordingly for VAE learning setup, which turns out to be more effective and numerically more stable than entropy-based regularization adopted in the BVI.
%
%Although a mixture model can be estimated using end-to-end gradient ascent methods, we demonstrate that our recursive estimation is more effective and less susceptible to problems of blind mixture estimation such as the collapsing of the mixture components. 
A crucial benefit of our approach is that the inference at test time requires a single feed-forward pass through the mixture inference network, making it significantly faster than the semi-amortized approaches. We show that our approach yields higher test data likelihood than the state-of-the-art on several benchmark datasets. 
\end{abstract}

%\vspace{-1.6em}

%%%%%%%%%%%%%%%%%%%%%%%%%%%%%%%%%%%%%%%%%%%%%%%%%%%%%%%%%%%%%%%%%%%%%%%%%%%%%%%
%%%%%%%%%%%%%%%%%%%%%%%%%%%%%%%%%%%%%%%%%%%%%%%%%%%%%%%%%%%%%%%%%%%%%%%%%%%%%%%
\section{Introduction}\label{sec:intro}

%\vspace{-0.5em}

Accurately modeling complex generative processes for high dimensional data (e.g., images) is a key task in deep learning. In many application fields, the Variational Autoencoder (VAE)~\cite{vae14,vae14r} was shown to be very effective for this task, endowed with the ability to interpret and directly control the latent variables that correspond to underlying hidden factors in data generation, a critical benefit over synthesis-only models such as GANs~\cite{gan}. The VAE adopts the %so-called 
{\em inference network} (aka encoder) that can perform %the 
test-time inference using a single feed-forward pass through a neural network. Although this feature, known as {\em amortized inference}, allows VAE to circumvent otherwise time-consuming procedures of solving the instance-wise variational optimization problem at test time, it %was observed that the amortized inference 
often results in inaccurate posterior approximation compared to the instance-wise variational optimization~\cite{cremer18}. 

Recently, semi-amortized approaches have been proposed to address this drawback. The main idea is to use an amortized encoder to produce a reasonable initial iterate, followed by instance-wise posterior fine tuning (e.g., a few gradient steps) to improve the posterior approximation~\cite{savae_ykim,savae_krishnan,savae_marino,vlae}. This is similar to the test-time model adaptation of the MAML~\cite{maml} in multi-task (meta) learning. However, this iterative gradient update may be computationally expensive during both training and test time: for training, some of the methods require Hessian-vector products for backpropagation, while at test time, one has to perform extra gradient steps for fine-tuning the variational optimization. Moreover, the performance of this approach is often very sensitive to the choice of the gradient step size and the number of gradient updates. % to achieve optimal performance-efficiency trade-off. 

In this paper, we consider a different approach; we build a mixture encoder model, for which we propose a recursive estimation algorithm that iteratively augments the current mixture with a new component encoder so as to reduce the divergence between the resulting variational and the true posteriors.  While the outcome is a (conditional) mixture inference model, which could also be estimated by end-to-end gradient descent~\cite{zobay}, our recursive estimation method is more effective and less susceptible to issues such as the mixture collapsing. %and dominating single component. 
This resiliency is attributed to our specific learning criteria for selecting a new mixture component: the new component has to improve the data likelihood (lower bound) and, at the same time, be as divergent as possible from the current mixture distribution, thus increasing the mixture diversity. 

%(WHY RECURSIVE LEARNING)
%Although a mixture model can be estimated using end-to-end gradient ascent methods, we demonstrate that our recursive estimation is more effective and less susceptible to problems of blind mixture estimation such as the collapsing of the mixture components. 

%(COMPARISON TO BOOSTED VI)
Although a recent family of methods called {\em Boosted Variational Inference} (BVI)~\cite{bvi_guo,bvi_locatello_nips,bvi_locatello_aistats,universal_bvi,bvi_adams} tackles this problem in a seemingly similar manner, our approach differs from BVI in several aspects.  Most notably, we address the recursive inference in VAEs in the form of amortized inference, while BVI is developed within the standard VI framework, leading to a non-amortized single optimization instance, inappropriate for VAEs in which the decoder also needs to be simultaneously learned. %\hl{if we say it's inappropriate, we should also say why}. 
%To address potential divergence of the projected gradient method of~\cite{bvi_locatello_aistats}, \cite{bvi_locatello_nips} emerged...
Furthermore, for the regularization strategy, required in the new component learning stage to avoid degenerate solutions, %that indefinitely increases the divergence from the current mixture
we employ the %{\em bounded KL barrier loss} 
{\em bounded KL loss}
instead of the previously used %{\em negative entropy penalization}, 
entropy regularization. 
This approach is better suited for amortized inference network learning in VAEs, %\hl{why?}, 
more effective 
%\hl{how? what do you mean by more effective?} 
as well as %empirically more 
numerically more stable than BVI (Sec.~\ref{sec:optim} for detailed discussions). %\hl{what? BVI?}.

Another crucial benefit of our approach is that the inference at test time is accomplished using a single feed-forward pass through the mixture inference network, a significantly faster process than the inference in semi-amortized methods. 
We show that our approach empirically yields higher test data likelihood than standard (amortized) VAE, existing semi-amortized approaches, and even the high-capacity flow-based encoder models on several benchmark datasets.

%\vspace{-0.2em}
%%%%%%%%%%%%%%%%%%%%%%%%%%%%%%%%%%%%%%%%%%%%%%%%%%%%%%%%%%%%%%%%%%%%%%%%%%%%%%%
%%%%%%%%%%%%%%%%%%%%%%%%%%%%%%%%%%%%%%%%%%%%%%%%%%%%%%%%%%%%%%%%%%%%%%%%%%%%%%%
\section{Background}\label{sec:background}

%\vspace{-0.7em}
%\vspace{-0.3em}

We denote by ${\bf x}$ observation (e.g., image) that follows the unknown distribution $p_{d}({\bf x})$. We aim to learn the VAE model that fits the given iid data $\{{\bf x}^i\}_{i=1}^N$ sampled from $p_d({\bf x})$. 
%The variational autoencoder (VAE)~\cite{vae14,vae14r}, a deep latent variable model, is one of the most popular models for this purpose. 
Specifically, letting ${\bf z}$ %$(\in \mathbb{R}^p)$ 
be the underlying %$p$-dimensional 
latent vector, the VAE %model 
is composed of a prior $p({\bf z}) = \mathcal{N}({\bf z}; {\bf 0}, {\bf I})$ and the conditional model $p_{\bm{\theta}}({\bf x}|{\bf z})$ where the latter, also referred to as the {\em decoder}, is defined as a tractable density (e.g., Gaussian) %or Bernoulli) 
whose parameters are the outputs of a deep network with weight parameters $\bm{\theta}$.

To fit the model, we aim to maximize the data log-likelihood, $\sum_{i=1}^N \log p_{\bm{\theta}}({\bf x}^i)$ where $p_{\bm{\theta}}({\bf x}) = \mathbb{E}_{p({\bf z)}}[p_{\bm{\theta}}({\bf x}|{\bf z})]$. %But optimizing or evaluating this marginal log-likelihood exactly is infeasible, due to the non-analytic form of the posterior $p_{\bm{\theta}}({\bf z}|{\bf x}) \propto p({\bf z}) p_{\bm{\theta}}({\bf x}|{\bf z})$. 
As evaluating the marginal likelihood exactly is infeasible, 
the variational inference %is an approximate learning method that 
aims to approximate the posterior by a density in some tractable family, that is, $p_{\bm{\theta}}({\bf z}|{\bf x}) \approx q_{\bm{\lambda}}({\bf z}|{\bf x})$ where $q_{\bm{\lambda}}({\bf z}|{\bf x})$ is a tractable density (e.g., Gaussian) with parameters $\bm{\lambda}$. For instance, if the Gaussian family is adopted, then $q_{\bm{\lambda}}({\bf z}|{\bf x}) = \mathcal{N}({\bf z}; \bm{\mu}, \bm{\Sigma})$, where $\{ \bm{\mu}, \bm{\Sigma} \}$ constitutes $\bm{\lambda}$. The approximate posterior $q_{\bm{\lambda}}({\bf z}|{\bf x})$ is often called the {\em encoder}. It is well known that the marginal log-likelihood is lower-bounded by the so-called {\em evidence lower bound} (ELBO, denoted by $\mathcal{L}$), 
%%%%
% \begin{align}
% & \log p_{\bm{\theta}}({\bf x}) \geq 
% \mathcal{L}(\bm{\lambda},\bm{\theta}; {\bf x}) := \label{eq:logpx_elbo} \\ 
% & \ \ \ \ \ \ \mathbb{E}_{q_{\bm{\lambda}}({\bf z}|{\bf x})} \big[ \log p_{\bm{\theta}}({\bf x}|{\bf z}) + \log p({\bf z}) - \log q_{\bm{\lambda}}({\bf z}|{\bf x}) \big], \nonumber
% \label{eq:logpx_elbo}
% \end{align}
%%
\begin{equation}
\log p_{\bm{\theta}}({\bf x}) \geq 
\mathcal{L}(\bm{\lambda},\bm{\theta}; {\bf x}) := \mathbb{E}_{q_{\bm{\lambda}}({\bf z}|{\bf x})} \big[ \log p_{\bm{\theta}}({\bf x}|{\bf z}) + \log p({\bf z}) - \log q_{\bm{\lambda}}({\bf z}|{\bf x}) \big], \label{eq:logpx_elbo}
\end{equation}
%%%%
where the gap in (\ref{eq:logpx_elbo}) is exactly the posterior approximation error $\textrm{KL}( q_{\bm{\lambda}}({\bf z}|{\bf x}) || p_{\bm{\theta}}({\bf z}|{\bf x}) )$. 

Hence, maximizing $\mathcal{L}(\bm{\lambda},\bm{\theta}; {\bf x})$ with respect to $\bm{\lambda}$ for the current $\bm{\theta}$ and the given input instance ${\bf x}$, amounts to finding the density in the variational family that best approximates the true posterior $p_{\bm{\theta}}({\bf z}|{\bf x})$. However, notice that the optimum $\bm{\lambda}$ must be specific to (i.e., dependent on) the input ${\bf x}$, and for some other input point ${\bf x}'$ one should do the ELBO optimization again to find the optimal encoder parameter $\bm{\lambda}'$ that approximates the posterior $p_{\bm{\theta}}({\bf z}|{\bf x}')$. The stochastic variational inference (SVI)~\cite{svi} directly implements this idea, and the approximate posterior inference for a new input point ${\bf x}$ in SVI amounts to solving the ELBO optimization on the fly by gradient ascent.

However, the downside is computational overhead since we have to perform iterative gradient ascent to have approximate posterior $q_{\bm{\lambda}}({\bf z}|{\bf x})$ for a new input ${\bf x}$. To remedy this issue, one can instead consider an ideal function $\bm{\lambda}^*({\bf x})$ that maps each input ${\bf x}$ to the optimal solution $\arg\max_{\bm{\lambda}} \mathcal{L}(\bm{\lambda},\bm{\theta}; {\bf x})$. We then introduce a deep neural network $\bm{\lambda}({\bf x}; \bm{\phi})$ with the weight parameters $\bm{\phi}$ as a universal function approximator of $\bm{\lambda}^*({\bf x})$. Then the ELBO, now denoted as $\mathcal{L}(\bm{\phi},\bm{\theta}; {\bf x})$, is optimized with respect to $\bm{\phi}$. This approach, called the {\em amortized} variational inference (AVI), was proposed in the original VAE~\cite{vae14}. A clear benefit of it is the computational speedup thanks to the feed-forward passing $\bm{\lambda}({\bf x}; \bm{\phi})$ used to perform posterior inference for a new input ${\bf x}$. 

Although AVI is computationally more attractive, it is observed that the quality of data fitting is degraded due to the amortization error, defined as an approximation error originating from the difference between $\bm{\lambda}^*({\bf x})$ and $\bm{\lambda}({\bf x}; \bm{\phi})$~\cite{cremer18}. That is, the AVI's computational advantage comes at the expense of reduced approximation accuracy; the SVI posterior approximation can be more accurate since we minimize the posterior approximation error $\textrm{KL}( q_{\bm{\lambda}}({\bf z}|{\bf x}) || p_{\bm{\theta}}({\bf z}|{\bf x}) )$ {\em individually} for each input ${\bf x}$. 
To address this drawback, the {\em semi-amortized} variational inference (SAVI) approaches have been proposed in~\cite{savae_ykim,savae_marino,savae_krishnan}. The main idea is to use the amortized encoder to produce a reasonably good initial iterate for the subsequent SVI optimization. 
The parameters $\bm{\phi}$ of the amortized encoder are trained in such a way that several steps of warm-start SVI gradient ascent would yield reduction of the instance-wise posterior approximation error, which is similar in nature to the gradient-based meta learning~\cite{maml} aimed at fast adaptation of the model to a new task in the multi-task meta learning.

However, the iterative gradient update procedure in SAVI is computationally expensive during both training and test times. For training, it requires backpropagation for the objective that involves gradients, implying the need for Hessian
evaluation (albeit finite difference approximation). %\footnote{To reduce computational cost, %overhead, %there were methods to approximate 
%the Hessian-vector products can be approximated by finite differences~\cite{hessian_lecun,hessian_domke}. %, also adopted in~\cite{savae_ykim}.
%} 
More critically, at test time, the inference requires a time-consuming gradient ascent optimization.  
Moreover, its performance is often quite sensitive to the choice of the gradient step size and the number of gradient updates; and it is difficult to tune these parameters to achieve optimal performance-efficiency trade-off. 
Although more recent work~\cite{vlae} mitigated the issue of choosing the step size by the first-order approximate solution method with the Laplace approximation, such linearization of the deep decoder network restricts its applicability to the models containing only fully connected layers, and makes it difficult to be applied to more structured models such as convolutional networks.

%\vspace{-0.3em}
%%%%%%%%%%%%%%%%%%%%%%%%%%%%%%%%%%%%%%%%%%%%%%%%%%%%%%%%%%%%%%%%%%%%%%%%%%%%%%%
%%%%%%%%%%%%%%%%%%%%%%%%%%%%%%%%%%%%%%%%%%%%%%%%%%%%%%%%%%%%%%%%%%%%%%%%%%%%%%%
\section{Recursive Mixture Inference Model (Proposed Method)}\label{sec:main}

\vspace{-0.7em}

%\hl{would be better to give it a catchy name instead of Our Method; maybe use RME as you use later?} 
Our method is motivated by the premise of the semi-amortized inference (SAVI), i.e., refining the variational posterior %of the amortized inference (AVI) 
to further reduce the difference from the true posterior. However, instead of doing the direct SVI gradient ascent as in SAVI, we introduce another amortized encoder model that augments the first amortized encoder to reduce the posterior approximation error. 

Formally, let $q_{\bm{\phi}}({\bf z}|{\bf x})$ be our amortized encoder model\footnote{This is a shorthand for $q_{\bm{\lambda}({\bf x}; \bm{\phi})}({\bf z}|{\bf x})$. We often drop the subscript and use $q({\bf z}|{\bf x})$  %${\bm{\phi}}$  
%q_{\bm{\phi}}({\bf z}|{\bf x})$ or 
for simplicity in notation.
} with the parameters $\bm{\phi}$. For the current decoder $\bm{\theta}$, % and the encoder $q({\bf z}|{\bf x})$, 
the posterior approximation error $\textrm{KL}( q({\bf z}|{\bf x}) || p_{\bm{\theta}}({\bf z}|{\bf x}) )$ equals -$\mathcal{L}(q,\bm{\theta}; {\bf x})$ (up to constant).\footnote{We often abuse the notation, either $\mathcal{L}(\bm{\phi},\bm{\theta}; {\bf x})$ or $\mathcal{L}(q,\bm{\theta}; {\bf x})$ interchangeably.} %, where the latter is the abbreviation for -$\mathcal{L}(\bm{\phi},\bm{\theta}; {\bf x})$. 
The goal is to find another amortized encoder model $q'({\bf z}|{\bf x})$ with the parameters ${\bm{\phi}'}$ such that, when convexly combined with $q({\bf z}|{\bf x})$ in a mixture $\epsilon q' + (1-\epsilon) q$ for some small $\epsilon>0$, the resulting {\em reduction of the posterior approximation error}, $\Delta\textrm{KL} := \mathcal{L}(\epsilon q' + (1-\epsilon) q, \bm{\theta}; {\bf x}) - \mathcal{L}(q, \bm{\theta}; {\bf x})$, is maximized.
%
%Note that $\Delta\textrm{KL}$ reduces to $0$ for the choice $q'=q$, suggesting no progress, hence this is not desired.  %\hl{Why is this statement necessary? One does not want to minimize the difference.} 
%We seek $q'$ (i.e., the amortized encoder parameters $\bm{\phi}'$) that maximizes $\Delta\textrm{KL}$, hence $q' \neq q$ while also expecting the optimal $q'$ to be highly distinct from $q$, unless $q$ is already near optimal. 
That is, we seek $\bm{\phi}'$ that maximizes $\Delta\textrm{KL}$.

%%%%%%%%

%\textbf{SAVI vs.~our method.}
\textbf{Compared to SAVI.}
The added encoder $q'$ can be seen as the means for correcting $q$, to reduce the mismatch between $q$ and the true %posterior 
$p_{\bm{\theta}}({\bf z}|{\bf x})$. In SAVI, this correction is done by explicit gradient ascent (finetuning) along $\bm{\phi}$ for every inference query, at train or test time, which is computationally expensive. In contrast, %in our approach 
we learn a differential amortized encoder at training time, which is fixed at test time, %and hence requires 
requiring only a single neural network feed-forward pass to obtain the approximate posterior. %\hl{check this}

%%%%%%%%

This encoder correction-by-augmentation can continue by regarding the mixture $\epsilon q' + (1-\epsilon) q$ as our current inference model to which another new amortized encoder will be added, with the recursion repeated a few times. This leads to a {\em mixture} model for the encoder,  $Q({\bf z}|{\bf x}) = \alpha_0 q({\bf z}|{\bf x}) + \alpha_1 q'({\bf z}|{\bf x}) + \cdots$, where $\sum_m \alpha_m = 1$. % are the mixing proportions. 
The main question is how to find the next encoder model to augment the current mixture $Q$. We do this by the functional gradient approach~\cite{funcgrad_friedman,funcgrad_mason}.

%%%%%%%%

\begin{figure}[t!]
\vspace{-1.5em}
\begin{center}
%
%\begin{subfigure}[b]{0.9\textwidth}
\centering
\includegraphics[trim = 3mm 3mm 0mm 2mm, clip, scale=0.276]{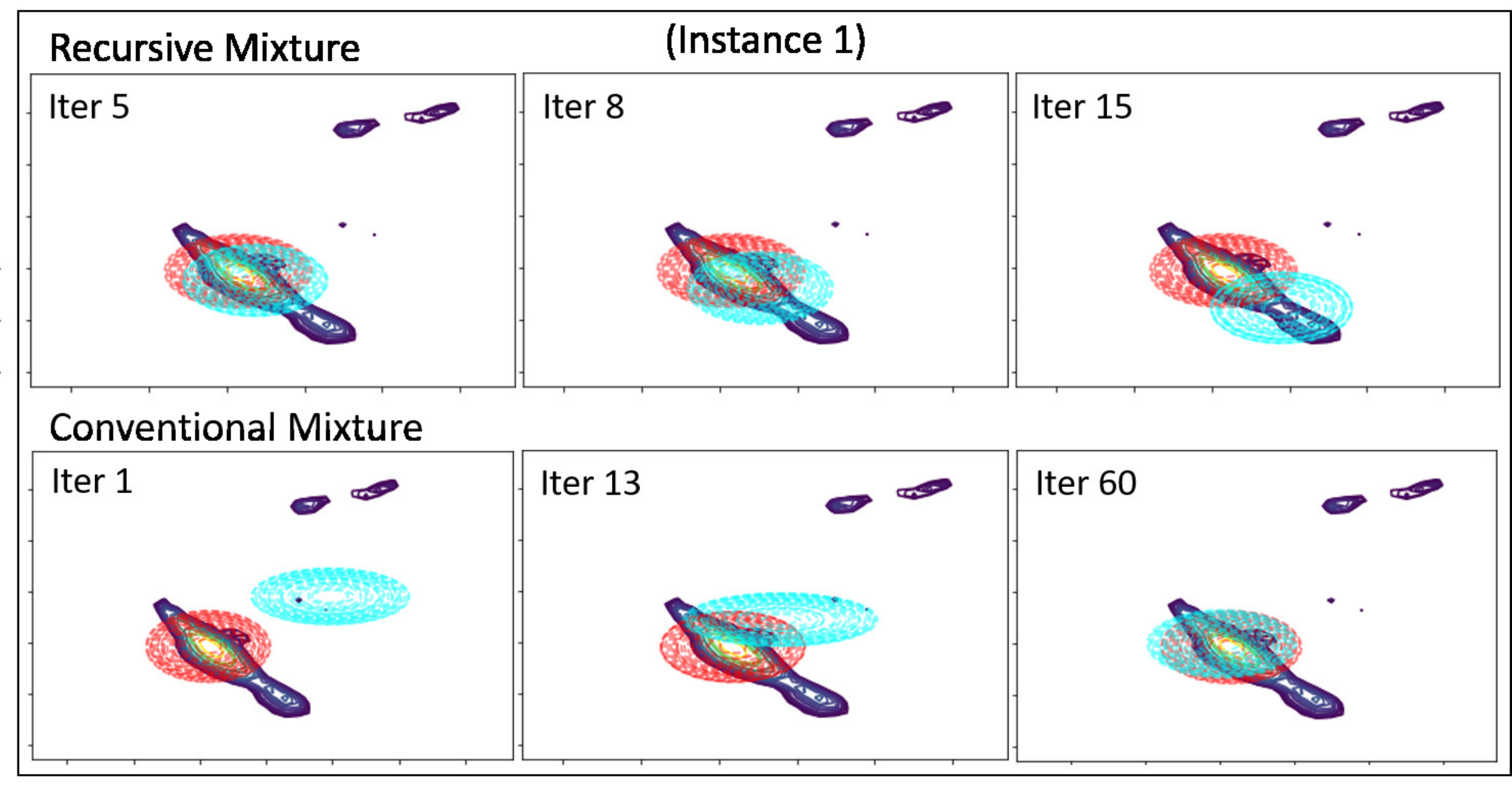} \ \ 
\includegraphics[trim = 3mm 3mm 0mm 2mm, clip, scale=0.276]{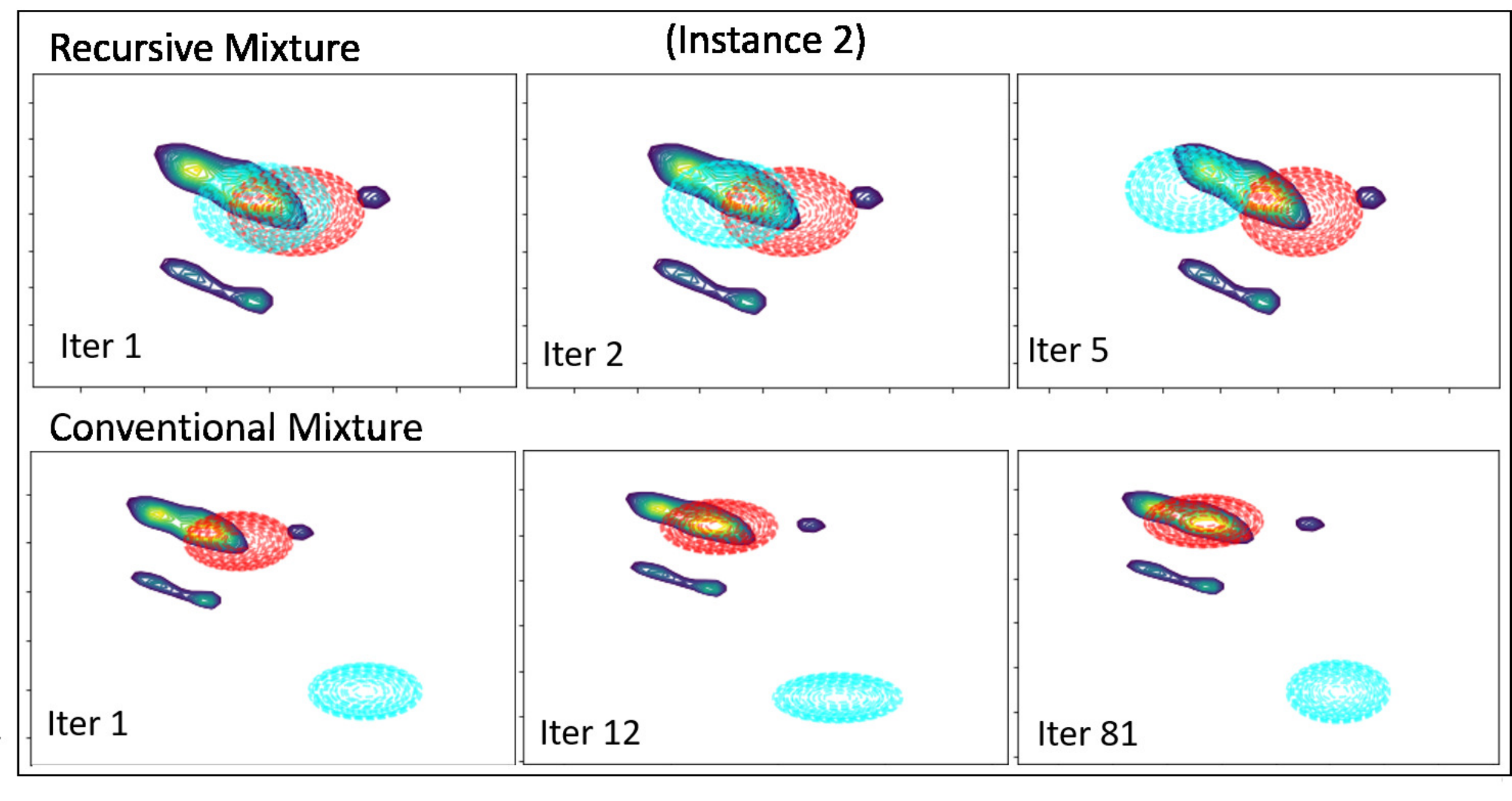}
\end{center}
\vspace{-1.2em}
\caption{
Illustration %of the learning progress of the recursive mixture estimation 
on MNIST using 2D latent ${\bf z}$ space. Results on two data instances (left and right) are shown. 
(\textbf{Top}) Our recursive estimation: The progress of learning the second mixture component is shown from left to right. The contour shows the true posterior $p({\bf z}|{\bf x})$, the red is %the first encoder component 
$q_0({\bf z}|{\bf x})$, the cyan is the second component that we learn here $q_1({\bf z}|{\bf x})$. We only trained $q_1$; remaining parameters (of the decoder and $q_0$) are fixed. %, i.e., we perform (\ref{eq:upd_phi1}) and (\ref{eq:upd_eta1}) only. 
Parameters of $q_1$ are initialized to those of $q_0$. 
(\textbf{Bottom}) Conventional (blind) mixture estimation by end-to-end gradient ascent. % on the same data instances. 
%The two components (Gaussians) are shown as red and cyan. We show the learning progress of the mixture encoder model while the decoder is fixed. 
For the instance 1 (left), the two components collapse onto each other. For the second (right), a single component (red) becomes dominant while the other (cyan) stays away, unutilized, from the support of the true posterior. %\hl{I suppose this is due to initialization?  May need to be careful, as one can claim a more sophisticated initialization tech could have been used.}
The cyan is initialized randomly to be different from the red (otherwise, it constitutes a local minimum). 
}
\vspace{-0.2em}
\label{fig:mnist_2d}
\end{figure}

\textbf{Functional gradients for mixture component search.} Following the functional gradient framework~\cite{funcgrad_friedman,funcgrad_mason}, the (ELBO) objective for the mixture $Q({\bf z}|{\bf x})$ can be expressed as a {\em functional}, namely a function that takes a density function $Q$ as input, 
%%%%
\begin{equation}
\vspace{-0.0em}
J(Q) := \mathbb{E}_{Q({\bf z}|{\bf x})} \big[ \log p_{\bm{\theta}}({\bf x}|{\bf z}) + \log p({\bf z}) - \log Q({\bf z}|{\bf x}) \big].
\label{eq:functional_obj}
\vspace{-0.0em}
\end{equation}
%%%%
%Suppose that 
Let $Q({\bf z}|{\bf x})$ be %is 
our current mixture. We aim to find %a new encoder %model 
$q({\bf z}|{\bf x})$ to be added to $Q$ by convex combination, 
%%%%
\begin{equation}
\vspace{-0.0em}
Q({\bf z}|{\bf x}) \leftarrow \epsilon q({\bf z}|{\bf x}) + (1-\epsilon) Q({\bf z}|{\bf x})
\label{eq:functional_update}
\vspace{-0.0em}
\end{equation}
%%%%
%$\epsilon q + (1-\epsilon) Q$ 
for some small $\epsilon>0$, that maximizes our objective functional $J$. 
To this end we take the functional gradient of the objective $J(Q)$ with respect to $Q$. For a given input ${\bf x}$, %this can be done by regarding 
we regard the function $Q({\bf z}|{\bf x})$ as an infinite-dimensional vector indexed by ${\bf z}$, and take the partial derivative at each ${\bf z}$, which yields:
%%%%
\begin{equation}
\vspace{-0.0em}
\frac{\partial J(Q)} {\partial Q({\bf z}|{\bf x})} = \log p_{\bm{\theta}}({\bf x}|{\bf z}) + \log p({\bf z}) - \log Q({\bf z}|{\bf x}) - 1.
\label{eq:functional_grad}
\vspace{-0.0em}
\end{equation}
%%%%
Since we have a convex combination (\ref{eq:functional_update}), the steepest ascent direction (\ref{eq:functional_grad}) needs to be projected onto the feasible function space $\{q(\cdot|{\bf x})-Q(\cdot|{\bf x}): q \in \mathcal{Q} \}$ where $\mathcal{Q} = \{ q_{\bm{\phi}} \}_{\bm{\phi}}$ is the set of variational densities realizable by the parameters $\bm{\phi}$. %In turn, %essence, 
Formally we solve the following optimization: % problem:
%%%%
\begin{equation}
\vspace{-0.0em}
\max_{q\in \mathcal{Q}} \ \bigg\langle q(\cdot|{\bf x}) - Q(\cdot|{\bf x}), \ \frac{\partial J(Q)} {\partial Q(\cdot|{\bf x})} \bigg\rangle,
\label{eq:functional_optim}
\vspace{-0.0em}
\end{equation}
%%%%
where $\langle \cdot, \cdot \rangle$ denotes the inner product in the function space. Using (\ref{eq:functional_grad}), and considering all training samples ${\bf x}\sim p_d({\bf x})$, the optimization (\ref{eq:functional_optim}) can be %equivalently 
written as:
%%%%
\begin{equation}
\vspace{-0.0em}
\max_{\bm{\phi}} \ \mathbb{E}_{%{\bf x} \sim 
p_d({\bf x})} 
\Big[ \mathbb{E}_{q_{\bm{\phi}}({\bf z}|{\bf x})} \big[ \log p_{\bm{\theta}}({\bf x}|{\bf z}) + \log p({\bf z}) - \log Q({\bf z}|{\bf x}) \big] \Big],
\label{eq:functional_optim_ultim}
\vspace{-0.0em}
\end{equation}
%%%%
%where $\mathbb{E}_{\bf x}[\cdot]$ denotes
where the outer expectation is with respect to the data distribution $p_d({\bf x})$. 
By adding and subtracting $\log q_{\bm{\phi}}({\bf z}|{\bf x})$ to and from the objective, we see that (\ref{eq:functional_optim_ultim}) can be rephrased as follows:
%%%%
\begin{equation}
\vspace{-0.0em}
\max_{\bm{\phi}} \ \mathbb{E}_{p_d({\bf x})} 
\Big[ \mathcal{L}(\bm{\phi},\bm{\theta}; {\bf x}) + 
\textrm{KL}(q_{\bm{\phi}}({\bf z}|{\bf x}) || Q({\bf z}|{\bf x})) \Big].
\label{eq:functional_optim_ultim2}
\vspace{-0.0em}
\end{equation}
%%%%
Note that (\ref{eq:functional_optim_ultim2}) gives us very intuitive criteria of how the new encoder component $q_{\bm{\phi}}$ should be selected: it has to maximize the ELBO (the first objective term), and at the same time, $q_{\bm{\phi}}$ should be {\em different} from the current mixture $Q$ (the KL term). That is, our next encoder has to keep explaining the data well (by large ELBO) while increasing the diversity of the encoder distribution (by large KL),  concentrating on those regions of the latent space that were poorly represented by the current %model 
$Q$. %In other words, the new component complements the regions not covered by the support of the current mixture. 
This supports our original %intention and 
intuition stated at the beginning of this section. See \autoref{fig:mnist_2d} for the illustration.

%%%%%%%%
%\vspace{-0.3em}
\textbf{Why recursive estimation.} Although we eventually form a (conditional) mixture model for the variational encoder, and such a mixture model can be estimated by end-to-end gradient descent, our recursive estimation is efficient and less susceptible to the known issues of blind mixture estimation, including collapsed mixture components and domination by a single component. 
This resiliency is attributed to our specific learning criteria for selecting a new mixture component: %the new component has to 
improve the data likelihood and at the same time be as distinct as possible from the current mixture,  %distribution, 
thus increasing diversity. 
See \autoref{fig:mnist_2d} for an illustrative comparison between our recursive and blind mixture estimation.

%%%%
\subsection{Optimization Strategy}\label{sec:optim}

\vspace{-0.5em} 

%\textbf{Optimization strategy.} 
Although we discussed the key idea of recursive mixture estimation, that is, at each step, fixing the current mixture $Q$ and add a new component $q$, it should be noted that the previously added components $q$'s (and their mixing proportions) need to be refined every time we update the decoder parameters $\bm{\theta}$. This is due to the VAE framework in which we have to learn the decoder in conjunction with the inference model, one of the main differences from the previous BVI approaches (See \autoref{sec:related}). %(\autoref{sec:boosted_vi}). 

\vspace{-0.3em}
To this end, we consider a mixture model $Q$ that consists of the {\em fixed} number ($M$) of components added to the initial component (denoted by $q_0$), namely
%%%%
\begin{equation}
\vspace{-0.0em}
Q({\bf z}|{\bf x}) = \alpha_0({\bf x}) q_0({\bf z}|{\bf x}) + \sum_{m=1}^M \alpha_m({\bf x}) q_m({\bf z}|{\bf x}),
\label{eq:mixture}
\vspace{-0.0em}
\end{equation}
%%%%
where $q_m({\bf z}|{\bf x})$ ($m=0,\dots,M$) are all amortized encoders whose parameters are denoted by $\bm{\phi}_m$, and $\alpha_m$ are the mixing proportions. Since the impact of each component can be different from instance to instance, we %define them as 
consider functions $\alpha_m({\bf x})$, instead of scalars. %Moreover, to 
To respect the idea of recursively adding %mixture 
components (i.e., $q_m$ with $\epsilon_m$), %it is easy to see that 
the mixing proportions conform to the following implicit structure: 
\vspace{-0.5em}
%%%%
\begin{equation}
\vspace{-0.0em}
\alpha_m({\bf x}) = \epsilon_{m}({\bf x}) \prod_{j=m+1}^M (1-\epsilon_j({\bf x})) \ \ \textrm{for} \ m=0,1,\dots,M \ \ (\textrm{let} \ \epsilon_{0}({\bf x}) = 1).
\vspace{-0.0em}
\end{equation}
%%%%
%\vspace{-0.1em}
This is %easily 
derived from the recursion, $Q_m = (1-\epsilon_m) Q_{m-1} + \epsilon_m q_m$ for $m=1,\dots,M$, where we denote by $Q_m$ the mixture formed by $q_0, q_1,\dots, q_m$ with $\epsilon_0 (=1)$,  $\epsilon_1,\dots,\epsilon_m$, and $Q_0 := q_0$. Hence $Q_M = Q$. 
Note also that we model $\epsilon_m({\bf x})$ as neural networks $\epsilon_m({\bf x}; \bm{\eta}_m)$ with %the weight 
parameters $\bm{\eta}_m$. % that are to be learned. %m=1,\dots,M$, which we learn. 
%To keep a single component from taking the majority of the density, we set an upper bound $\epsilon_{\max}$ for the network output, which is done by applying a final softmax layer (Supplement for details). 
% So, we have $\epsilon_m({\bf x}; \bm{\eta}_m)$ where there can be network weight sharing across $m=1,\dots,M$.
%$\epsilon({\bf x})$, and accordingly  in  (\ref{eq:mixture}).   
\begin{comment}
%%%%
\begin{eqnarray}
\alpha_M({\bf x}) &=& \epsilon_M({\bf x}) \\
\alpha_{M-1}({\bf x}) &=&  \epsilon_{M-1}({\bf x})(1-\epsilon_M({\bf x})) \\
&\;\;\vdots \notag \\
%&\vdots& \\
\alpha_1({\bf x}) &=& \epsilon_{1}({\bf x}) %\prod_{m=2}^M (1-\epsilon_m({\bf x})) 
(1-\epsilon_2({\bf x})) \cdots (1-\epsilon_M({\bf x})) \ \ \ \ \
\\
\alpha_0({\bf x}) &=& %\prod_{m=1}^M (1-\epsilon_m({\bf x})).
(1-\epsilon_1({\bf x})) \cdots (1-\epsilon_M({\bf x}))
\end{eqnarray}
%%%%
\end{comment}

\vspace{-0.2em}
Now we describe our recursive mixture learning algorithm. 
As we seek to update all components simultaneously together with the decoder $\bm{\theta}$, we employ gradient ascent optimization with {\em all parameters iteratively and repeatedly}. Our algorithm is described in Alg.~\ref{alg:main}. 
Notice that for the ${\bm \phi}$ update in the algorithm, we used the BKL which stands for {\em Bounded KL}, in place of KL. The KL term in (\ref{eq:functional_optim_ultim2}) is to be {\em maximized}, and it can be easily unbounded; In typical situations, $\textrm{KL}(q||Q)$ can become arbitrarily large by having $q$ concentrate on the region where $Q$ has zero support. To this end, we impose an upper barrier on the KL term, that is,  
$\textrm{BKL}(q || Q) = \max(C, \textrm{KL}(q || Q))$, so that increasing KL beyond the barrier point $C$ gives no incentive. %The $C$ is a positive user-specified parameter (e.g., 
$C=500.0$ works well empirically. 

\vspace{-0.2em}
Similar degeneracy issues have been dealt with in the previous BVI approaches for non-VAE variational inference~\cite{bvi_guo,bvi_locatello_nips}. Most approaches attempted to regularize small entropy when optimizing the new components to be added. 
%(penalizing negative entropy when optimizing the newly added component $q_m$), 
However, the entropy regularization may be less effective for the iterative refinement of the mixture components within the VAE framework, since we have indirect control of the component models (and their entropy values) only through the density parameter networks $\bm{\lambda}({\bf x}; \bm{\phi})$ in $q_{\bm{\lambda}({\bf x}; \bm{\phi})}({\bf z}|{\bf x})$ (i.e., amortized inference). Furthermore, it encourages the component densities to have large entropy all the time as a side effect, which can lead to a suboptimal solution in certain situations. Our upper barrier method, on the other hand, regularizes the component density only if they are too close (within the range of $C$ KL divergence) to the current mixture, rendering it better chance to find an optimal 
solution outside the $C$-ball of the current mixture. In fact, the empirical results in \autoref{sec:comp_bvi} demonstrate that our strategy %bounded KL barrier loss strategy %of directly upper bounding the KL 
leads to better performance.

\vspace{-0.2em}
The nested %training 
loops in Alg.~\ref{alg:main} may appear computationally costly, however, the outer loop usually takes a few epochs (usually no more than $20$) since we initialize all components $q_m$ identically with the trained encoder parameters of the standard VAE (afterwards, the components quickly move away from each other due to the BKL term). The mixture order $M$ (the number of the inner iterations) is typically small as well (e.g., between 1 and 4), which renders the algorithm fairly efficient in practice. 

% More specifically, the update equation for $\bm{\phi}_m$ ($m=1,\dots,M$) is modified as:
% %%%%
% \begin{equation}
% \bm{\phi}_m: \ \max_{\bm{\phi}_m} \ \mathbb{E}_{\bf x}%{p_d({\bf x})} 
% \Big[ \mathcal{L}(\bm{\phi}_m, \bm{\theta}; {\bf x}) -  %\textrm{KL}(q_{\bm{\phi}_M}({\bf z}|{\bf x}) || Q_{M-1}({\bf z}|{\bf x})) \Big]
% \big(C-\textrm{KL}(q_{\bm{\phi}_m} || Q_{m-1})\big)_+ \Big],
% \end{equation}
% %%%%
% where $(a)_+ = \max(0,a)$ and $C$ is a positive user-specified constant for the KL upper barrier (e.g., $C=500.0$ works well empirically). 

%The upper barrier method that we suggest can avoid it. % (i.e., there exists a region $X \subset \mathcal{X}$ such that $q(X)>0$ but $Q(X)=0$.)
 
%Note that the first step (\ref{eq:upd_phi0}) coincides with the encoder update formula for the conventional VAE.
%The above steps are performed %\st{simultaneously or} {\em sequentially}, repeated until convergence. 

\newcommand\inlineeqno{\stepcounter{equation}\ (\theequation)}
%%%%
\begin{algorithm}[t!]
%\vspace{-0.4em}
  \caption{Recursive Learning Algorithm for Mixture Inference Model.}
  \label{alg:main}
 \begin{small}
\begin{algorithmic}
   \STATE {\bfseries Input:} %Initial inference components $\{ q_m({\bf z}|{\bf x};\bm{\phi}_m) \}_{m=0}^M$, $\{\epsilon_m({\bf x}; \bm{\eta}_m) \}_{m=1}^M$ and decoder $p_{\bm{\theta}}({\bf x}|{\bf z})$. \STATE %{\bfseries hyper-parameters:} 
%\ \ \ \ \ \ \ \ \ \ \ \ Learning rate $\gamma$. KL upper bound $C$.
     Initial $\{ q_m({\bf z}|{\bf x};\bm{\phi}_m) \}_{m=0}^M$, $\{\epsilon_m({\bf x}; \bm{\eta}_m) \}_{m=1}^M$, and $p_{\bm{\theta}}({\bf x}|{\bf z})$. Learning rate $\gamma$. KL bound $C$.
   \STATE {\bfseries Output:} Learned inference and decoder models.
   \STATE {\bfseries Let:}
      $Q_m = (1-\epsilon_m) Q_{m-1} + \epsilon_m q_m$ %($1\leq m \leq M$) 
      ($m=1\dots M$), $Q_0=q_0$. 
      $\textrm{BKL}(p||q) = \max(C,\textrm{KL}(p||q))$.
      %$\textrm{BKL}(p||q) = -\big(C-\textrm{KL}(p||q)\big)_+$
   \REPEAT 
      \STATE Sample a batch of data ${\bf B}$ from $p_d({\bf x})$.
      \STATE Update $q_0({\bf z}|{\bf x};\bm{\phi}_0)$: \  
         $\bm{\phi}_0 \leftarrow \bm{\phi}_0 + \gamma \nabla_{\bm{\phi}_0} 
            \mathbb{E}_{{\bf x}\sim{\bf B}} \big[ \mathcal{L}(q_0, \bm{\theta}; {\bf x}) \big]$. %\inlineeqno$.
      \FOR{$m=1,\dots,M$}
         \STATE Update $q_m({\bf z}|{\bf x};\bm{\phi}_m)$: \ %$q_m$: \ %
            $\bm{\phi}_m \leftarrow \bm{\phi}_m + \gamma \nabla_{\bm{\phi}_m} 
               \mathbb{E}_{{\bf x}\sim{\bf B}} \big[ \mathcal{L}(q_m, \bm{\theta}; {\bf x}) 
                 + \textrm{BKL}(q_m || Q_{m-1}) \big]$.
         \STATE Update $\epsilon_m({\bf x}; \bm{\eta}_m)$: \ %\ \ \ \  $\epsilon_m$: \ %
            $\bm{\eta}_m \leftarrow \bm{\eta}_m + \gamma \nabla_{\bm{\eta}_m} 
               \mathbb{E}_{{\bf x}\sim{\bf B}} \big[ 
               \mathcal{L}\big((1-\epsilon_m) Q_{m-1} + \epsilon_m q_m, \bm{\theta}; {\bf x}\big) \big]$.
      \ENDFOR
    %   \STATE Update $q_1({\bf z}|{\bf x};\bm{\phi}_1)$: \ 
    %      $\bm{\phi}_1 \leftarrow \bm{\phi}_1 + \gamma \nabla_{\bm{\phi}_1} 
    %         \mathbb{E}_{{\bf x}\sim{\bf B}} \big[ \mathcal{L}(q_1, \bm{\theta}; {\bf x}) 
    %           + \textrm{BKL}(q_1 || Q_0) \big]$.
    %   %
    %   \STATE Update $\epsilon_1({\bf x}; \bm{\eta}_1)$: \ %\ \ \ \ 
    %      $\bm{\eta}_1 \leftarrow \bm{\eta}_1 + \gamma \nabla_{\bm{\eta}_1} 
    %         \mathbb{E}_{{\bf x}\sim{\bf B}} \big[ 
    %         \mathcal{L}((1-\epsilon_1) Q_{0} + \epsilon_1 q_1, \bm{\theta}; {\bf x}) \big]$.
    %   %
    %   \STATE Update $q_2({\bf z}|{\bf x};\bm{\phi}_2)$: \ 
    %      $\bm{\phi}_2 \leftarrow \bm{\phi}_2 + \gamma \nabla_{\bm{\phi}_2} 
    %         \mathbb{E}_{{\bf x}\sim{\bf B}} \big[ \mathcal{L}(q_2, \bm{\theta}; {\bf x}) 
    %           + \textrm{BKL}(q_2 || Q_1) \big]$.
    %   %
    %   \STATE Update $\epsilon_2({\bf x}; \bm{\eta}_2)$: \ %\ \ \ \ 
    %      $\bm{\eta}_2 \leftarrow \bm{\eta}_2 + \gamma \nabla_{\bm{\eta}_2} 
    %         \mathbb{E}_{{\bf x}\sim{\bf B}} \big[ 
    %         \mathcal{L}((1-\epsilon_2) Q_{1} + \epsilon_2 q_2, \bm{\theta}; {\bf x}) \big]$.
      %
      \STATE Update $p_{\bm{\theta}}({\bf x}|{\bf z})$: \ 
         $\bm{\theta} \leftarrow \bm{\theta} + \gamma \nabla_{\bm{\theta}} 
            \mathbb{E}_{{\bf x}\sim{\bf B}} \big[ \mathcal{L}(Q_M, \bm{\theta}; {\bf x}) \big]$.
   \UNTIL{convergence}
   %\STATE {\bfseries return} $Q = Q_M$.
\end{algorithmic}
\end{small}
%\vspace{-0.1em}
\end{algorithm}
\section{Related Work}\label{sec:related}

%\vspace{-1.0em}

%\textbf{Semi-amortized VAE.} 
%As discussed in \autoref{sec:background}, the 
The VAE's issue of amortization error was raised recently~\cite{cremer18}, and the semi-amortized inference approaches~\cite{savae_ykim,savae_marino,savae_krishnan} attempted to address the issue by performing the SVI gradient updates at test time.
%
%\textbf{Normalizing flows.} 
Alternatively one can enlarge the representational capacity of the encoder network, yet still amortized inference. %One of the most 
A popular approach is the flow-based models that apply nonlinear invertible transformations to VAE's variational posterior~\cite{hfvae,iafvae}. The transformations could be complex autoregressive mappings, while they can also model full covariance matrices via efficient parametrization % so as to increase flexibility
to represent arbitrary rotations, i.e., cross-dimensional dependency. % beyond the axis-parallel diagonal covariances in the VAE. 
%
%\textbf{Functional gradient.} 
Our use of functional gradient in designing a learning objective stems from the framework in~\cite{funcgrad_friedman,funcgrad_mason}. Mathematically elegant and flexible in the learning criteria, the framework was more recently exploited in~\cite{pfg} to unify seemingly different machine learning paradigms. %, including variational inference, adversarial learning, and reinforcement learning. 
%
%\textbf{Other mixture modeling.} 
%There were 
Several mixture-based approaches aimed to extend the representational capacity of the variational inference model. In~\cite{sivi} the variational parameters were mixed with a flexible distribution. %In~\cite{zobay} Gaussian-mixture approximation is used, while 
In~\cite{vampprior} the prior is modeled as a mixture (aggregate posterior), while~\cite{boovae} attempted to tighten the lower bound by matching optimal prior with functional Frank-Wolfe. 
%Another attempt to enrich the VAE's model complexity is the so called 
\begin{comment}
The hierarchical decompositional mixture~\cite{spvae} %. Although the model is termed a mixture, it 
is a hybrid model of VAE and the sum-product network~\cite{spn} (VAEs as leaf nodes), %leading to the model family 
very distinct from the framework that we focus on in this paper. %; hence, empirical comparison to this model is out of scope. 
\end{comment}

\vspace{-0.3em}
\textbf{Boosted VI.} Previously, there were approaches to boost the inference network in variational inference similar to our idea~\cite{bvi_guo,bvi_locatello_nips,bvi_locatello_aistats,universal_bvi,bvi_adams}, where some of them~\cite{bvi_locatello_nips,bvi_locatello_aistats,universal_bvi} focused on theoretical convergence analysis, inspired by the Frank-Wolfe~\cite{fw_revisit} interpretation of the greedy nature of the algorithm in the  infinite-dimensional (function) space. However, these approaches all aimed for stochastic VI in the non-VAE framework, hence non-amortized inference, whereas we consider amortized inference in the VAE framework in which both the decoder and the inference model need to be learned.  
%They used entropy regularization for the new component to avoid degenerate solutions...
%Despite similarity of the algorithms, our approaches differ from the boosted VI as discussed in \autoref{sec:boosted_vi}. 
We briefly summarize the main differences between the previous BVI approaches and ours as follows:
%%%%
%\begin{enumerate}
%
%\item %They are on SVI, ours on amortized VI for the VAE model. More specifically, they 
1) We learn $Q({\bf z}|{\bf x})$, a density functional of input ${\bf x}$, % where we learn the network (function) from $x$ to $z$. 
while BVI optimizes $Q({\bf z})$, a single variational density (not a function of ${\bf x}$), and thus involves only single optimization. %(eg, ${\bf z}$ is rather the parameters of the model in Bayesian modeling and they do posterior approximation), 
%
%\item %We optimize the inference model together with the decoder. % $p_{\bm{\theta}}({\bf x}|{\bf z})$. 
2) Within the VAE framework, as the decoder is not optimal in the course of training, we update the decoder and all the inference components iteratively and repeatedly. %This means that our whole procedure (\ref{eq:upd_phi0}--\ref{eq:upd_theta}) repeatedly. -- Put the entire algorithm for clarification. 
%
%\item 
3) To avoid degeneracy in KL maximization, we employ the bounded KL instead of BVI's entropy penalization, better suited for amortized inference %in VAEs 
and more effective in practice.
%The KL maximization term in the next component $q$ optimization part, the previous work employed the entropy regularization (i.e., penalizing small entropy $q$) to prevent it from approaching degenerate delta-like density. On the other hand, we rather use the KL barrier (not incentivizing for $q$ to be highly divergent from current $Q$). From empirical evaluations, we show that this KL barrier is more viable for amortized inference network optimization than the previous entropy regularization.  -- As the degenerate $q$ can have the high KL that can easily overtake the entropy penalty, these entropy regularization approaches are numerically less stable. On the other hand, ours have hard barrier on the KL (no incentive beyond a certain threshold value of KL), our model is far more stable numerically. Note that our $q$ is the amortized inference model, hence, we cannot directly enforcing (eg, in Guo et al.) $\log\det\bm{\Sigma}$ to be large, since $\bm{\Sigma}$ is not directly controllable but rather an output of a network...(?)
%
%\item 
4) The instant impacts of the components, $\epsilon({\bf x})$ are also modeled input-dependent (as neural networks) rather than tunable scalars as in BVI. 
%
%\end{enumerate}
%%%%

%\vspace{-1.0em}
%%%%%%%%%%%%%%%%%%%%%%%%%%%%%%%%%%%%%%%%%%%%%%%%%%%%%%%%%%%%%%%%%%%%%%%%%%%%%%%
%%%%%%%%%%%%%%%%%%%%%%%%%%%%%%%%%%%%%%%%%%%%%%%%%%%%%%%%%%%%%%%%%%%%%%%%%%%%%%%
\section{Evaluations}\label{sec:expmt}

%\vspace{-1.0em}

We test %demonstrate the superiority of 
the proposed recursive inference model\footnote{The code is publicly available from  \url{https://github.com/minyoungkim21/recmixvae}} %VAE model %over existing state-of-the-arts 
on several benchmark datasets. 
%\begin{comment}
We %particularly 
highlight %the 
improved test likelihood scores and %the 
reduced % test 
inference time, compared to %the
semi-amortized VAEs. 
%\end{comment}
We also contrast with %the 
flow models that aim to increase modeling accuracy using %controlled 
high capacity encoders. 

\vspace{-0.3em}
\textbf{Competing approaches.} \textbf{VAE}: The standard VAE model (amortized inference)~\cite{vae14,vae14r}.
\textbf{SA}: The semi-amortized VAE~\cite{savae_ykim}. We fix the SVI gradient step size as $10^{-3}$, but vary the number of SVI steps from $\{1, 2, 4, 8\}$. 
\textbf{IAF}: The autoregressive-based flow model for the encoder $q({\bf z}|{\bf x})$~\cite{iafvae}, which has richer expressiveness than VAE's %post-Gaussian 
Gaussian encoder. % adopted in VAE models. 
%The number of flows %(i.e., the number of compositions) 
%is chosen from $\{1, 2, 4, 8\}$. 
%
\textbf{HF}: The Householder flow encoder model that represents the full covariance using the Householder transformation~\cite{hfvae}. The numbers of flows for IAF and HF %(i.e., the number of compositions) 
are chosen from $\{1, 2, 4, 8\}$. 
\textbf{ME}: For a baseline comparison, % with our recursive mixture estimation, 
we also consider the same mixture encoder model, but unlike our recursive mixture learning, the model is trained conventionally, end-to-end; all mixture components' parameters are updated simultaneously. The number of mixture components is chosen from $\{2,3,4,5\}$.
\textbf{RME}: Our proposed recursive mixture encoder model. We  vary the number of %new components to be added 
additional components $M$ from $\{1,2,3,4\}$, leading to mixture order $2$ to $5$. %The first component $\bm{\phi}_0$ is 
All components are initialized identically with the VAE's encoder. %(and updated thereafter). 
%Unlike RME, only the first component of the ME is initialized with the VAE's encoder, and the rest chosen randomly, since initializing all components identically would constitute a local maximum, unable to be updated further. 
See Supplement for the details. % of the hyperparameters. 

\vspace{-0.3em}
\textbf{Datasets.} %We deal with %the following 
%five benchmark datasets: 
\textbf{MNIST}~\cite{mnist}, %$(28 \times 28 \times 1)$. 
\textbf{OMNIGLOT}~\cite{omniglot}, %: $24,345$ training images and $8,070$ test images where each image is of dimension $(28 \times 28 \times 1)$. We randomly hold out $10\%$ of the training set as a validation set. 
    %\item \textbf{FashionMNIST}~\cite{fashion_mnist}: $60,000$ training images and $10,000$ test images where each image is of dimension $(28 \times 28 \times 1)$. We randomly hold out $10\%$ of the training set as a validation set. 
%\textbf{CIFAR10}~\cite{cifar10}, %: $50,000$ training images and $10,000$ test images where each image is of dimension $(32 \times 32 \times 3)$. We randomly hold out $10\%$ of the training set as a validation set. 
\textbf{SVHN}~\cite{svhn}, % $73,257$ training images and $26,032$ test images where each image is of dimension $(32 \times 32 \times 3)$. We randomly hold out $10\%$ of the training set as a validation set. 
and \textbf{CelebA}~\cite{celeba}. %: $202,599$ tightly cropped face images of size $(64 \times 64 \times 3)$. 
%For the CelebA, we use tightly cropped face images of size $(64 \times 64 \times 3)$, and randomly split the data into $80\%/10\%/10\%$ train/validation/test sets. For the other datasets, we follow the train/test partitions provided in the data, where $10\%$ of the training sets are randomly held out as validation sets. 
%For CelebA, we use tightly cropped face images of size $(64 \times 64 \times 3)$, and randomly split data into $80\%/10\%/10\%$ train/validation/test sets. For others, we follow train/test partitions provided in the data, where $10\%$ of the training sets are randomly held out for validation. % sets. 
We follow train/test partitions provided in the data, where $10\%$ of the training sets are randomly held out for validation. For CelebA, we randomly split data into $80\%/10\%/10\%$ train/validation/test sets.

\vspace{-0.3em}
\textbf{Network architectures}. We adopt the convolutional neural networks for the encoder and decoder %\footnote{More precisely, we used the convolutional network for the encoder and the transposed convolution network for the decoder.} 
models for all competing approaches. This is because the convolutional networks are believed to outperform fully connected networks for many tasks in the image domain~\cite{cnn_imagenet,cnn_szegedy,cnn_dcgan}. We also provide empirical evidence in the Supplement by comparing the test likelihood performance between the two architectures.\footnote{
%In particular, it is shown (in the Supplement) that the fully-connected decoder architecture is inferior to the deconvnet decoder that we adopted, when the two architectures have roughly equal numbers of parameters. This is why we exclude comparison with the recent Laplacian approximation approach of~\cite{vlae} in the main paper.
Fully-connected decoder architectures are inferior to the deconvnet %that we adopted, 
when the number of parameters are roughly equal. % which makes us exclude comparison with the Laplacian approximation approach of~\cite{vlae}.  
%They use the first-order approximation solver method to obtain the mode of the true posterior, but such linearization of a deep network is only computationally feasible for {\em fully connected} decoder models. On the other hand, our recursive mixture learning admits arbitrary types of encoder/decoder architectures, which is another advantage. 
This is why we exclude comparison with the recent~\cite{vlae}, but see Supplement for the results. 
%In the Supplement, we report the performance of the Laplace approximation~\cite{vlae}. % with fully connected decoder models.
}
%
%For the encoder architecture, we first apply $L$ convolutional layers with $(4 \times 4)$-pixels kernels, followed by two fully-connected layers with hidden layers dimension $h$. For the decoder, the input images first go through two fully connected layers, followed by $L$ deconvolution\footnote{We used {\em transposed convolution} networks.} layers with $(4 \times 4)$-pixels filters. Here, $L=3$ for all datasets except the CelebA which has $L=4$, and  $h=256$ for the MNIST/OMNIGLOT and $h=512$ for the others. For the mixing proportion network $\epsilon({\bf x};\bm{\eta})$ in our recursive mixture learning method (RME), we used a fully connected network with one hidden layer of dimension $10$. For more details of the network architectures, refer to the Supplement. 
For the details of the network architectures, refer to the Supplement. 

\definecolor{lor}{rgb}{1,0.85,0}
\definecolor{or}{rgb}{1,0.60,0}
\definecolor{dor}{rgb}{1,0.20,0}
\newcommand\Tstrut{\rule{0pt}{2.2ex}}         % = `top' strut
\newcommand\Bstrut{\rule[-0.9ex]{0pt}{0pt}}   % = `bottom' strut
%%%% 
\begin{table}%[t]
\vspace{-1.5em}
\centering
\caption{Test log-likelihood scores %(unit in nat) 
estimated by IWAE sampling. %importance weighted sampling~\cite{iwae}. % with 100 samples. 
The %figures in the 
parentheses next to model names indicate: the number of SVI steps in SA, the number of flows in IAF and HF, and the %number of mixture components 
mixture order in ME and RME. The superscripts are the standard deviations. 
The best (on average) results are boldfaced in $\color{red} \textrm{\textbf{red}}$. 
In each column, the statistical significance of the difference between the best model (red) and each competing model, is depicted as color: anything non-colored indicates $p\leq 0.01$ %\textrm{black}$ 
(strongly distinguished), 
$p \in (0.01,0.05]$ %$0.01<p\leq 0.05 = 
as $\color{lor} \textrm{yellow-orange}$, 
$p \in (0.05,0.1]$ %$0.05<p\leq 0.10 = 
as $\color{or} \textrm{orange}$, 
$p>0.1$ as $\color{dor} \textrm{red orange}$ (little evidence of difference) by the  Wilcoxon signed rank test. Best viewed in color.
}
\vspace{-0.6em}
\label{tab:test_loglik_all}
%\vskip 0.05in
\begin{scriptsize}
%\begin{footnotesize}
%\begin{small}
%\begin{sc}
\centering
\scalebox{0.95}{
\begin{tabular}{l|ll|ll|ll|ll}
\toprule
%\multirow{2}{*}{}
Dataset & \multicolumn{2}{c|}{MNIST} & \multicolumn{2}{c|}{OMNIGLOT} &  \multicolumn{2}{c|}{SVHN} & \multicolumn{2}{c}{CelebA} 
\\ \cline{2-9}
\ \ $\textrm{dim}({\bf z})$ & \multicolumn{1}{|c}{$20$} & \multicolumn{1}{c|}{$50$} & \multicolumn{1}{|c}{$20$} & \multicolumn{1}{c|}{$50$} &  \multicolumn{1}{|c}{$20$} & \multicolumn{1}{c|}{$50$} & \multicolumn{1}{|c}{$20$} & \multicolumn{1}{c}{$50$} \\
\hline\hline
VAE\Tstrut & $930.7^{3.9}$ & $1185.7^{3.9}$ & $501.6^{1.6}$ & $801.6^{4.0}$ & $4054.5^{14.3}$ & $5363.7^{21.4}$ & $12116.4^{25.3}$ & $15251.9^{39.7}$ \\ \hline 
SA$^{(1)}$\Tstrut & $921.2^{2.3}$ & $1172.1^{1.8}$ & $499.3^{2.5}$ & $792.7^{7.9}$ & $4031.5^{19.0}$ & $5362.1^{35.7}$ & $12091.1^{21.6}$ & $15285.8^{29.4}$ \\
SA$^{(2)}$ & $932.0^{2.4}$ & $1176.3^{3.4}$ & $501.0^{2.7}$ & $793.1^{4.8}$ & $4041.5^{15.5}$ & $5377.0^{23.2}$ & $12087.1^{21.5}$ & $15252.7^{29.0}$ \\
SA$^{(4)}$ & $925.5^{2.6}$ & $1171.3^{3.5}$ & $488.2^{1.8}$ & $794.4^{1.9}$ & $4051.9^{22.2}$ & $5391.7^{20.4}$ & $12116.3^{20.5}$ & $15187.3^{27.9}$ \\
SA$^{(8)}$ & $928.1^{3.9}$ & $1183.2^{3.4}$ & $490.3^{2.8}$ & $799.4^{2.7}$ & $4041.6^{9.5}$ & $5370.8^{18.5}$ & $12100.6^{22.8}$ & $15096.5^{27.2}$ \\ \hline
IAF$^{(1)}$\Tstrut & $934.0^{3.3}$ & $1180.6^{2.7}$ & $489.9^{1.9}$ & $788.8^{4.1}$ & $4050.0^{9.4}$ & $5368.3^{11.5}$ & $12098.0^{20.6}$ & $15271.2^{28.6}$ \\
IAF$^{(2)}$ & $931.4^{3.7}$ & $1190.1^{1.9}$ & $494.9^{1.4}$ & $795.7^{2.7}$ & $4054.6^{10.5}$ & $5360.0^{10.0}$ & $12104.5^{21.8}$ & $15262.2^{27.8}$ \\
IAF$^{(4)}$ & $926.3^{2.6}$ & $1178.1^{1.6}$ & $496.0^{2.0}$ & $775.1^{2.2}$ & $4048.6^{8.7}$ & $5338.1^{10.2}$ & $12094.6^{22.6}$ & $15261.0^{28.1}$ \\
IAF$^{(8)}$ & $934.1^{2.4}$ & $1150.0^{2.2}$ & $498.8^{2.3}$ & $774.7^{2.9}$ & $4042.0^{9.6}$ & $5341.8^{10.1}$ & $12109.3^{22.0}$ & $15241.5^{27.9}$ \\ \hline
HF$^{(1)}$\Tstrut & $917.2^{2.6}$ & $\color{dor} 1204.3^{4.0}$ & $488.6^{2.0}$ & $795.9^{3.3}$ & $4028.8^{9.7}$ & $5372.0^{10.1}$ & $12077.2^{31.4}$ & $15240.5^{27.6}$ \\
HF$^{(2)}$ & $923.9^{3.1}$ & $1191.5^{10.8}$ & $495.9^{1.8}$ & $784.5^{4.8}$ & $4030.7^{9.9}$ & $5376.6^{10.2}$ & $12093.0^{25.6}$ & $15258.2^{30.3}$ \\
HF$^{(4)}$ & $927.3^{2.8}$ & $1197.2^{1.5}$ & $487.0^{2.7}$ & $799.7^{3.2}$ & $4038.4^{9.7}$ & $5371.8^{9.8}$ & $12082.0^{27.0}$ & $15266.5^{29.5}$ \\
HF$^{(8)}$ & $928.5^{3.1}$ & $1184.1^{1.8}$ & $488.3^{2.4}$ & $794.6^{4.0}$ & $4035.9^{8.9}$ & $5351.1^{11.1}$ & $12087.3^{25.5}$ & $15248.7^{29.7}$ \\ \hline
ME$^{(2)}$\Tstrut & $926.7^{3.0}$ & $1152.8^{1.7}$ & $491.7^{1.4}$ & $793.4^{3.8}$ & $4037.2^{11.0}$ & $5343.2^{13.1}$ & $12072.7^{23.3}$ & $15290.5^{29.3}$ \\
ME$^{(3)}$ & $933.1^{4.1}$ & $1162.8^{4.7}$ & $491.2^{2.1}$ & $807.5^{4.9}$ & $4053.8^{16.1}$ & $5367.7^{15.8}$ & $12100.3^{21.7}$ & $15294.6^{28.3}$ \\
ME$^{(4)}$ & $914.7^{2.3}$ & $\color{red} {\bf 1205.1}^{2.3}$ & $491.3^{1.8}$ & $732.0^{3.1}$ & $4061.3^{12.0}$ & $5191.9^{18.5}$ & $12092.2^{22.6}$ & $15270.7^{20.6}$ \\
ME$^{(5)}$ & $920.6^{1.9}$ & $1198.5^{3.5}$ & $478.0^{2.8}$ & $805.7^{3.8}$ & $4057.5^{12.2}$ & $5209.2^{12.8}$ & $12095.3^{25.1}$ & $15268.8^{27.5}$ \\ \hline
RME$^{(2)}$\Tstrut & $\color{lor} 943.9^{1.6}$ & $\color{or} 1201.7^{0.9}$ & $\color{dor} 508.2^{1.2}$ & $\color{red} {\bf 821.0}^{3.1}$ & $\color{dor} 4085.3^{9.7}$ & $\color{or} 5403.2^{10.2}$ & $\color{dor} 12193.1^{23.5}$ & $\color{dor} 15363.0^{31.7}$ \\
RME$^{(3)}$ & $\color{dor} 945.1^{1.6}$ & $\color{or} 1202.4^{1.0}$ & $\color{dor} 507.5^{1.1}$ & $\color{dor} 820.4^{0.9}$ & $\color{dor} 4085.9^{9.8}$ & $\color{dor} 5405.1^{10.4}$ & $\color{dor} 12192.3^{23.5}$ & $\color{dor} 15365.6^{31.4}$ \\
RME$^{(4)}$ & $\color{red} {\bf 945.2}^{1.6}$ & $\color{dor} 1203.1^{1.0}$ & $\color{dor} 509.0^{1.2}$ & $\color{dor} 819.9^{0.9}$ & $\color{dor} 4080.7^{9.9}$ & $\color{or} 5403.8^{10.2}$ & $\color{dor} 12192.6^{23.4}$ & $\color{dor} 15364.3^{31.5}$ \\
RME$^{(5)}$ & $\color{dor} 945.0^{1.7}$ & $\color{dor} 1203.7^{1.0}$ & $\color{red} {\bf 509.1}^{1.4}$ & $\color{dor} 819.9^{0.9}$ & $\color{red} {\bf 4086.9}^{10.9}$ & $\color{red} {\bf 5405.5}^{8.5}$ & $\color{red} {\bf 12194.2}^{11.5}$ & $\color{red} {\bf 15366.2}^{12.7}$ \\
\bottomrule
\end{tabular}
}
\vspace{-0.8em}
%\end{sc}
%\end{footnotesize}
\end{scriptsize}
%\end{small}
\vspace{-0.6em}
\end{table}
\setlength{\intextsep}{5pt}%
\setlength{\columnsep}{5pt}%
\begin{wraptable}[28]{r}{0.22\columnwidth}
%\begin{table}%[t!]
\vspace{-0.8em}
\centering
\setlength{\FrameSep}{2pt}
\caption{%(Per-batch) 
Test data log-likelihood scores for the {\bf Binary MNIST}. Our results are in the column titled ``CNN''. The column ``FC'' is excerpted from~\cite{vlae}. % (Table 2).  
}
\label{tab:binary_mnist}
%\vskip 0.05in
%\begin{small}
\begin{scriptsize}
\begin{sc}
\centering
\setlength{\tabcolsep}{2pt}
\begin{tabular}{lcc}
\toprule
%& M & O & Ci & S & Ce \\
& CNN & FC \\
\midrule
VAE & -84.49 & -85.38 \\ \hline
SA$^{(1)}$ & -83.64 & -85.20 \\
SA$^{(2)}$ & -83.79 & -85.10 \\
SA$^{(4)}$ & -83.85 & -85.43 \\
SA$^{(8)}$ & -84.02 & -85.24 \\ \hline
IAF$^{(1)}$ & -83.37 & -84.26 \\
IAF$^{(2)}$ & -83.15 & -84.16 \\
IAF$^{(4)}$ & -83.08 & -84.03 \\
IAF$^{(8)}$ & -83.12 & -83.80 \\ \hline
HF$^{(1)}$ & -83.82 & -85.27 \\
HF$^{(2)}$ & -83.70 & -85.31 \\
HF$^{(4)}$ & -83.87 & -85.22 \\
HF$^{(8)}$ & -83.76 & -85.41 \\ \hline
ME$^{(2)}$ & -83.77 & - \\
ME$^{(3)}$ & -83.81 & - \\
ME$^{(4)}$ & -83.83 & - \\
ME$^{(5)}$ & -83.75 & - \\ \hline 
VLAE$^{(2)}$ & - & -83.72 \\
VLAE$^{(3)}$ & - & -83.84 \\
VLAE$^{(4)}$ & - & -83.73 \\
VLAE$^{(5)}$ & - & -83.60 \\ \hline 
RME$^{(2)}$ & -83.14 & - \\
RME$^{(3)}$ & -83.14 & - \\
RME$^{(4)}$ & -83.09 & - \\
RME$^{(5)}$ & -83.15 & - \\ 
\bottomrule
\end{tabular}
\end{sc}
\end{scriptsize}
%\end{small}
%\end{tcolorbox}
%\end{table}
%\vspace{-1em}
\end{wraptable}
%%%%
%%%%%%%%
\vspace{-0.3em}
\textbf{Experimental setup}. We vary the latent %dimension 
$\textrm{dim}({\bf z})$, %In the previous work~\cite{cremer18,vlae}, they used fully connected networks for encoder/decoder with two different complexities, small and large. But it turns out that convnet significantly outperforms the full-con nets provided that the number of parameters of the two architectures are (roughly) equal. Instead, what matters more is the latent dimension, hence we vary the $\textrm{dim}({\bf z})$ from small to large. 
%either 
small (20) or large (50).\footnote{The results for $\textrm{dim}({\bf z})=10$ and $100$, also on the \textbf{CIFAR10} dataset~\cite{cifar10}, are reported in the Supplement.} %chosen from $\{10,20,50,100\}$. 
To report the test log-likelihood scores $\log p({\bf x})$, we use the importance weighted sampling estimation (IWAE) method~\cite{iwae} %. More specifically, $\textrm{IWAE} = \log \frac{1}{K} \sum_{i=1}^K 
%  \frac{ p({\bf x},{\bf z}_i) } { q({\bf z}_i|{\bf x}) }$,
%where ${\bf z}_1,\dots,{\bf z}_K$ are i.i.d.~samples from $q({\bf z}|{\bf x})$. %It can be shown that 
%IWAE lower bounds $\log p({\bf x})$ and can be arbitrarily close to the target as the number of samples $K$ grows. We use $K=100$ throughout the experiments. 
with 100 samples (Supplement for details). 
For each model/dataset, we perform 10 runs with different random train/validation splits, where each run consists of three trainings by starting with different random model parameters, among which only one model with the best validation result is chosen. 
%We also performed the one-sided Wilcoxon signed rank test for every pair (the best model, non-best model), using the 10 log-likelihood scores per model. 
%we repeat the training procedure three times with different random seeds, and report the best results. 
%(\hl{Do mention in the text that we ran 10 runs by random train/val/test splits where each run we have three runs with different random seeds, the best chosen.})  

\vspace{-0.3em}

%%%%%%%%%%%%%%%%%%%%%%%%%%%%%%%%%%%%%%%%%%%%%%%%%%%%%%%%%%%%%%%%%%%%%%%%%%%%%%%
\subsection{Results}\label{sec:results}

\vspace{-0.7em} 

The test log-likelihood scores are summarized in Table~\ref{tab:test_loglik_all}.\footnote{%For the MNIST results, the test log-likelihood scores of the competing methods 
The MNIST results mismatch those reported in the related work (e.g.,~\cite{vampprior}). Significantly higher scores. This is because we adopt the Gaussian decoder models, not the binary decoders, for all competing methods.} 
%\autoref{tab:mnist} (MNIST)\footnote{For the MNIST results, the test log-likelihood scores of the competing methods mismatch those reported in the related work (e.g.,~\cite{vampprior}). Significantly higher scores. This is because we adopt the Gaussian decoder models, not the binary decoders, for all competing methods.}, \autoref{tab:omniglot} (OMNIGLOT), \autoref{tab:cifar10} (CIFAR10), \autoref{tab:svhn} (SVHN), and \autoref{tab:celeba} (CelebA). %The qualitative results of reconstructed and synthesized images can be found in the Supplement. 
% Analysis/interpretation for each dataset...
Overall the results indicate 
that our recursive mixture encoder (RME) outperforms the competing approaches consistently for all datasets. %We provide interpretation of these results below.
To see the statistical significance, 
we performed the one-sided Wilcoxon signed rank test for every pair (the best model, non-best model). The results indicate that this superiority is statistically significant. %using the 10 log-likelihood scores per model. 

\vspace{-0.1em}
\textbf{Comparison to ME.} %The blind end-to-end mixture learning (ME) is sensitive to the initial parameters. 
With one exception, specifically ME (4) with $\textrm{dim}({\bf z}) = 50$ on the MNIST, the blind end-to-end mixture learning (ME) consistently underperforms our RME. As also illustrated in \autoref{fig:mnist_2d}, the blind mixture estimation can potentially suffer from mixture collapsing and single dominant component issues. The fact that even the VAE often performs comparably to the ME with different mixture orders supports this observation. 
On the other hand, our recursive mixture estimation is more robust to the initial parameters. Due to its incremental learning nature, it "knows" the regions in the latent space ill-represented by the current mixture, then updates %learns and adds new 
mixture components to complement those regions. This %sophisticated 
strategy %in selecting new mixture components 
allows the RME to effectively model highly multi-modal posterior distributions, yielding more robust and accurate variational posterior approximation. % for the deep generative model. 

\vspace{-0.1em}
\textbf{Comparison to SA.} The semi-amortized approach (SA) sometimes achieves improvement over the VAE, but not consistently. In particular, its performance is generally very sensitive to the number of SVI gradient update steps. This is another drawback of the SA, where the gradient-based adaption has to be performed at the test time. Although one could adjust the gradient step size (in place of currently used fixed step size) to improve the performance,  there is little principled way to tune the step size at test time that can attain optimal accuracy and inference time trade off. The number of SVI steps in the SA may correspond to the mixture order %(the number of components to be added) 
in our RME model, and the results show that increasing the mixture order usually improves, and not deteriorate, the generalization performance. 

%minor refinement from the VAE's amortized inference. 
%This highlights the issue with semi-amortized approaches that are strongly sensitive to the gradient step size and the number of update steps.

\vspace{-0.1em}
\textbf{Comparison to IAF/HF.} %These flow-based models basically apply nonlinear invertible transformations to VAE's post-Gaussian variational posterior, so as to construct more complex densities. The IAF adopts the autoregressive flows, being capable of representing highly nonlinear non-Gaussian conditional densities (perhaps subsuming the true posteriors), and the HF aims to model full covariance matrices for the variational posterior, increasing flexibility to represent arbitrary rotations (cross-dimensional dependencies) beyond the axis-parallel diagonal covariances in the VAE. 
Although flow models have rich representational capacity, possibly with full covariance matrices (HF), % (perhaps subsuming the true posteriors), 
the improvement over the VAE is limited compared to our RME; the models sometimes perform not any better than the VAE. 
%and tend to %possibly overfit 
The failure of the flow-based models may originate from the difficulty of optimizing the complex encoder models. (Similar observations were made in related previous work~\cite{vlae}). This result signifies that sophisticated and discriminative learning criteria are critical, beyond just enlarging the structural capacity of the neural networks, %an evidence similarly observed from the failure of the conventional mixture estimation.
similarly observed from the failure of conventional mixtures.

\vspace{-0.1em}
\textbf{Non-Gaussian likelihood model.}
Our empirical evaluations were predominantly conducted with the convolutional architectures on real-valued image data. For the performance of our model with non-convolutional (fully connected) network architectures, the readers can refer to Table 5 and 6 in the supplementary material. 
For the binarized input images, we have conducted extra experiments on the {\bf Binary MNIST} dataset. The binary images can be modeled by a Bernoulli likelihood in the decoder. Table~\ref{tab:binary_mnist} summarized the results. We have set the latent dimension $\dim({\bf z})=50$, 
and used the same CNN architectures as before, except that the decoder output is changed from Gaussian to Bernoulli. We also include the reported results from~\cite{vlae} for comparison, 
which  employed the same latent dimension $50$ and fully connected encoder/decoder networks with similar model complexity as our CNNs'. 
%Due to lack of time, we only report mean scores averaged over three runs. 
As shown, IAF and our RME performs equally the best, although the performance differences among the competing approaches are not very pronounced compared to real-valued image cases.

%%%%%%%%%%%%%%%%%%%%%%%%%%%%%%%%%%%%%%%%%%%%%%%%%%%%%%%%%%%%%%%%%%%%%%%%%%%%%%%
\subsection{Test Inference Time}\label{sec:inf_time}

\vspace{-0.5em} 

Another key advantage of our recursive mixture inference is the computational efficiency of test-time inference, comparable to that of VAE. Unlike the semi-amortized approaches, where one performs the SVI gradient adaptation at test time, the inference in our RME is merely a single feed forward pass through our mixture encoder network. That is, once training is done, our mixture inference model remains fixed, with no adaptation required. 

% %%%%
% \begin{table}%[t]
% \centering
% \caption{(Per-batch) Test inference time (milliseconds) with batch size 128. %The latent dimension 
% $\textrm{dim}({\bf z})=50$. %MNIST & OMNIGLOT & CIFAR10 & SVHN & CelebA
% }
% \label{tab:inf_time}
% %\vskip 0.05in
% %\begin{small}
% \begin{scriptsize}
% \begin{sc}
% \centering
% \begin{tabular}{lccccc}
% \toprule
% %& M & O & Ci & S & Ce \\
% & \scriptsize{MNIST} & \scriptsize{OMNIG.} & \scriptsize{CIFAR10} & \scriptsize{SVHN} & \scriptsize{CelebA} \\
% \midrule
% VAE & \ \ 3.6 & \ \ 4.8 & \ \ 3.7 & \ \ 2.2 & \ \ 2.7 \\ \hline
% %
% SA ($1$) & \ \ 9.7 & 11.6 & \ \ 9.8 & \ \ 7.0 & \ \ 8.4 \\
% SA ($2$) & 18.1 & 19.2 & 16.8 & 15.5 & 13.8 \\
% SA ($4$) & 32.2 & 34.4 & 27.9 & 30.1 & 27.1 \\
% SA ($8$) & 60.8 & 65.7 & 60.5 & 60.3 & 53.8 \\ \hline
% %
% IAF ($1$) & \ \ 4.8 & \ \ 5.7 & \ \ 5.1 & \ \ 3.4 & \ \ 4.4 \\
% IAF ($2$) & \ \ 5.9 & \ \ 6.4 & \ \ 5.6 & \ \ 3.7 & \ \ 5.1 \\
% IAF ($4$) & \ \ 6.2 & \ \ 7.0 & \ \ 6.3 & \ \ 4.7 & \ \ 5.7 \\
% IAF ($8$) & \ \ 7.7 & \ \ 8.2 & \ \ 7.6 & \ \ 5.7 & \ \ 7.7 \\ \hline
% %
% RME ($2$) & \ \ 4.7 & \ \ 5.4 & \ \ 4.9 & \ \ 3.2 & \ \ 4.2 \\
% RME ($3$) & \ \ 4.9 & \ \ 5.5 & \ \ 5.1 & \ \ 3.6 & \ \ 4.1 \\
% RME ($4$) & \ \ 4.6 & \ \ 5.3 & \ \ 5.1 & \ \ 3.5 & \ \ 4.2 \\
% RME ($5$) & \ \ 4.8 & \ \ 5.6 & \ \ 5.1 & \ \ 3.3 & \ \ 4.8 \\ 
% \bottomrule
% \end{tabular}
% \end{sc}
% \end{scriptsize}
% %\end{small}
% \end{table}
% %%%%
%

To verify this empirically, we measure the actual inference time for the  competing approaches. The per-batch test inference times (batch size 128) on all benchmark datasets are shown in \autoref{tab:inf_time}. 
%%%%
\setlength{\intextsep}{5pt}%
\setlength{\columnsep}{5pt}%
\begin{wraptable}[14]{r}{0.38\columnwidth}
%\begin{table}%[t!]
\vspace{-1.0em}
\centering
\setlength{\FrameSep}{2pt}
\caption{%(Per-batch) 
%Test 
Inference time (milliseconds). % with batch size 128. %The latent dimension 
%$\textrm{dim}({\bf z})=50$. %MNIST & OMNIGLOT & CIFAR10 & SVHN & CelebA
}
\label{tab:inf_time}
%\vskip 0.05in
%\begin{small}
\begin{scriptsize}
\begin{sc}
\centering
\setlength{\tabcolsep}{2pt}
\begin{tabular}{lcccc}
\toprule
%& M & O & Ci & S & Ce \\
& \scriptsize{MNIST} & \scriptsize{OMNIG.} & \scriptsize{SVHN} & \scriptsize{CelebA} \\
\midrule
VAE & \ \ 3.6 & \ \ 4.8 & \ \ 2.2 & \ \ 2.7 \\ \hline
SA$^{(1)}$ & \ \ 9.7 & 11.6 & \ \ 7.0 & \ \ 8.4 \\
SA$^{(2)}$ & 18.1 & 19.2 & 15.5 & 13.8 \\
SA$^{(4)}$ & 32.2 & 34.4 & 30.1 & 27.1 \\
SA$^{(8)}$ & 60.8 & 65.7 & 60.3 & 53.8 \\ \hline
IAF$^{(1)}$ & \ \ 4.8 & \ \ 5.7 & \ \ 3.4 & \ \ 4.4 \\
IAF$^{(2)}$ & \ \ 5.9 & \ \ 6.4 & \ \ 3.7 & \ \ 5.1 \\
IAF$^{(4)}$ & \ \ 6.2 & \ \ 7.0 & \ \ 4.7 & \ \ 5.7 \\
IAF$^{(8)}$ & \ \ 7.7 & \ \ 8.2 & \ \ 5.7 & \ \ 7.7 \\ \hline
RME$^{(2)}$ & \ \ 4.7 & \ \ 5.4 & \ \ 3.2 & \ \ 4.2 \\
RME$^{(3)}$ & \ \ 4.9 & \ \ 5.5 & \ \ 3.6 & \ \ 4.1 \\
RME$^{(4)}$ & \ \ 4.6 & \ \ 5.3 & \ \ 3.5 & \ \ 4.2 \\
RME$^{(5)}$ & \ \ 4.8 & \ \ 5.6 & \ \ 3.3 & \ \ 4.8 \\ 
\bottomrule
\end{tabular}
\end{sc}
\end{scriptsize}
%\end{small}
%\end{tcolorbox}
%\end{table}
%\vspace{-1em}
\end{wraptable}
%%%%
To report the results, for each method and each dataset, we run the inference over the entire test set batches, measure the running time, then take the per-batch average. We repeat the procedure five times and report the average. All models are run on the same machine with a single GPU (RTX 2080 Ti), Core i7 3.50GHz CPU, and 128 GB RAM. While we only report test times for $\textrm{dim}({\bf z})=50$, the impact of the latent dimension appears to be less significant.
%We only report test times for the latent dimension $\textrm{dim}({\bf z})=50$ %(the highest dimension in our experimental setup) as the impact of the latent dimension appears to be less significant. 

As expected, the semi-amortized approach suffers from the computational overhead of test-time gradient updates, with the inference time significantly increased as the number of updates increases. Our RME is comparable to VAE, and faster than IAF (with more than a single flow), which verifies our claim. Interestingly, increasing the mixture order in our model rarely affects the inference time, due to intrinsic parallelization of the feed forward pass through the multiple mixture components networks, leading to inference time as fast as that of VAE. %the single component model (VAE). %, where the number of mixture components do not really matter. 

%%%%%%%%%%%%%%%%%%%%%%%%%%%%%%%%%%%%%%%%%%%%%%%%%%%%%%%%%%%%%%%%%%%%%%%%%%%%%%%
\subsection{Comparison with Boosted VI's Entropy Regularization}\label{sec:comp_bvi}

\vspace{-0.5em} 

Recall that our RME adopted the bounded KL (BKL) loss to avoid degeneracy in the component update stages. Previous boosted VI (BVI) approaches employ different regularization, namely penalizing small entropy for the new components. However, such indirect regularization  can be less effective for the iterative refinement of the mixture components within the VAE framework (the second last paragraph of \autoref{sec:optim}). To verify this claim, we test our RME models with the BKL loss replaced by the BVI's entropy regularization. 
More specifically, %there are two schemes: \textbf{BVI}$^\dagger$: \cite{bvi_guo}'s Gaussian entropy based regularization, that is, penalizing small $\log\det\bm{\Sigma}$ where $\bm{\Sigma}$ is the (diagonal) covariance of the new component $q({\bf z}|{\bf x})$, and % to be updated. 
%\textbf{BVI}$^\diamond$: 
following the scheme of~\cite{bvi_locatello_nips}, we replace our BKL loss by $\nu \cdot \mathbb{E}_{q({\bf z}|{\bf x})}[-\log q({\bf z}|{\bf x})]$ estimated by Monte Carlo, 
%regularization of the negative entropy of $q({\bf z}|{\bf x})$ whose 
where $\nu = 1/\sqrt{t+1}$ %\frac{1}{\sqrt{t+1}}$ 
is the impact that decreases as the training iteration $t$.\footnote{We also tested a slight variant, \cite{bvi_guo}'s closed-form Gaussian entropy $\log\det\bm{\Sigma}$ where $\bm{\Sigma}$ is the (diagonal) covariance of the new component $q({\bf z}|{\bf x})$. The results were very similar to the scheme of~\cite{bvi_locatello_nips}. See Supplement.}
See \autoref{tab:kl_reg} for the results. 
This empirical result %in \autoref{sec:comp_bvi} 
demonstrates that our bounded KL loss consistently yields better performance than entropy regularization. We also observe that our BKL loss leads to numerically more stable solutions: For entropy regularization, we had to reduce the learning rate to the tenth of that of BKL to avoid NaNs. 

%%%% 
\begin{table}%[b!]
\vspace{-1.5em}
\centering
\caption{Comparison with the BVI's entropy regularization~\cite{bvi_locatello_nips}. The same color scheme as \autoref{tab:test_loglik_all}.
%\textbf{BVI}$^\dagger$: \cite{bvi_guo}'s Gaussian entropy based regularization (i.e., penalizing small $\log\det\bm{\Sigma}$ where $\bm{\Sigma}$ is the (diagonal) covariance matrix of the new component $q({\bf z}|{\bf x})$ to be optimized. 
%\textbf{BVI}$^\diamond$: \cite{bvi_locatello_nips}'s regularization of the negative entropy of $q({\bf z}|{\bf x})$ whose impact decreases $\frac{1}{\sqrt{t+1}}$ as a function of training iteration $t$, as suggested.
}
\vspace{-0.6em}
\label{tab:kl_reg}
%\vskip 0.05in
\begin{scriptsize}
%\begin{footnotesize}
%\begin{small}
%\begin{sc}
\centering
\scalebox{0.95}{
\begin{tabular}{l|ll|ll|ll|ll}
\toprule
%\multirow{2}{*}{}
Dataset & \multicolumn{2}{c|}{MNIST} & \multicolumn{2}{c|}{OMNIGLOT} & \multicolumn{2}{c|}{SVHN} & \multicolumn{2}{c}{CelebA} 
\\ \cline{2-9}
\ \ $\textrm{dim}({\bf z})$ & \multicolumn{1}{|c}{$20$} & \multicolumn{1}{c|}{$50$} &  \multicolumn{1}{|c}{$20$} & \multicolumn{1}{c|}{$50$} &  \multicolumn{1}{|c}{$20$} & \multicolumn{1}{c|}{$50$} &  \multicolumn{1}{|c}{$20$} & \multicolumn{1}{c}{$50$} \\
\hline\hline
RME$^{(2)}$\Tstrut & $\color{lor} 943.9^{1.6}$ & $1201.7^{0.9}$ & $\color{dor} 508.2^{1.2}$ & $\color{red} {\bf 821.0}^{3.1}$ & $\color{dor} 4085.3^{9.7}$ & $\color{or} 5403.2^{10.2}$ & $\color{dor} 12193.1^{23.5}$ & $\color{dor} 15363.0^{31.7}$ \\
RME$^{(3)}$ & $\color{dor} 945.1^{1.6}$ & $\color{lor} 1202.4^{1.0}$ & $\color{dor} 507.5^{1.1}$ & $\color{dor} 820.4^{0.9}$ & $\color{dor} 4085.9^{9.8}$ & $\color{dor} 5405.1^{10.4}$ & $\color{dor} 12192.3^{23.5}$ & $\color{dor} 15365.6^{31.4}$ \\
RME$^{(4)}$ & $\color{red} {\bf 945.2}^{1.6}$ & $\color{lor} 1203.1^{1.0}$ & $\color{dor} 509.0^{1.2}$ & $\color{dor} 819.9^{0.9}$ & $\color{dor} 4080.7^{9.9}$ & $\color{or} 5403.8^{10.2}$ & $\color{dor} 12192.6^{23.4}$ & $\color{dor} 15364.3^{31.5}$ \\
RME$^{(5)}$ & $\color{dor} 945.0^{1.7}$ & $\color{red} {\bf 1203.7}^{1.0}$ & $\color{red} {\bf 509.1}^{1.4}$ & $\color{dor} 819.9^{0.9}$ & $\color{red} {\bf 4086.9}^{10.9}$ & $\color{red} {\bf 5405.5}^{8.5}$ & $\color{red} {\bf 12194.2}^{11.5}$ & $\color{red} {\bf 15366.2}^{12.7}$ \\
\hline
% RME$^{(2)}$\Tstrut & 943.9 & 1201.7 & 508.2 & $\color{red} {\bf 821.0}$ & 4085.3 & 5403.2 & 12193.1 & 15363.0 \\
% RME$^{(3)}$ & 945.1 & 1202.4 & 507.5 & 820.4 & 4085.9 & 5405.1 & 12192.3 & 15365.6 \\
% RME$^{(4)}$ & $\color{red} {\bf 945.2}$ & 1203.1 & 509.0 & 819.9 & 4080.7 & 5403.8 & 12192.6 & 15364.3 \\
% RME$^{(5)}$ & 945.0 & $\color{red} {\bf 1203.7}$ & $\color{red} {\bf 509.1}$ & 819.9 & $\color{red} {\bf 4086.9}$ & $\color{red} {\bf 5405.5}$ & $\color{red} {\bf 12194.2}$ & $\color{red} {\bf 15366.2}$ \\
% %
% \hline%\hline
%BVI$^\dagger$ ($2$) & 939.7 & 1189.6 & 507.8 & 817.1 & 2094.4 & 2775.8 & 4077.3 & 5388.3 & 12133.6 & 15207.3 \\
%BVI$^\dagger$ ($3$) & 939.4 & 1192.1 & 507.8 & 816.6 & 2094.7 & 2776.2 & 4076.7 & 5383.9 & 12146.6 & 15249.6 \\
%BVI$^\dagger$ ($4$) & 937.6 & 1191.5 & 507.8 & 816.9 & 2095.0 & 2776.5 & 4073.2 & 5371.3 & 12128.7 & 15084.9 \\
%BVI$^\dagger$ ($5$) & 931.7 & 1181.7 & 508.1 & 816.4 & 2095.1 & 2776.8 & 4071.2 & 5377.7 & 12087.5 & 15051.7 \\ \hline
%
%BVI$^\diamond$ 
BVI$^{(2)}$\Tstrut & $939.7^{2.8}$ & $1196.2^{2.8}$ & $\color{lor} 507.9^{2.2}$ & $\color{lor} 817.1^{3.3}$ & $\color{lor} 4077.3^{10.3}$ & $5388.2^{10.2}$ & $12133.5^{25.1}$ & $15206.4^{28.2}$ \\
%BVI$^\diamond$ 
BVI$^{(3)}$ & $939.5^{2.9}$ & $1191.6^{2.9}$ & $\color{or} 507.8^{2.2}$ & $\color{lor} 816.6^{3.4}$ & $\color{lor} 4076.6^{10.3}$ & $5384.2^{10.5}$ & $12146.5^{22.4}$ & $15249.5^{28.1}$ \\
%BVI$^\diamond$ 
BVI$^{(4)}$ & $937.8^{2.9}$ & $1191.6^{2.8}$ & $\color{dor} 507.8^{2.3}$ & $\color{lor} 816.8^{3.4}$ & $\color{lor} 4073.1^{10.2}$ & $5371.1^{10.4}$ & $12127.7^{22.3}$ & $15085.8^{28.4}$ \\
%BVI$^\diamond$ 
BVI$^{(5)}$ & $931.2^{3.0}$ & $1183.1^{2.9}$ & $\color{dor} 508.2^{2.3}$ & $\color{lor} 816.4^{3.3}$ & $\color{lor} 4071.2^{10.2}$ & $5378.1^{10.1}$ & $12092.3^{22.3}$ & $15052.5^{28.0}$ \\ % \hline
\bottomrule
\end{tabular}
}
%\end{sc}
%\end{footnotesize}
\end{scriptsize}
%\end{small}
\vspace{-1.0em}
\end{table}
%%%%

%%%%%%%%%%%%%%%%%%%%%%%%%%%%%%%%%%%%%%%%%%%%%%%%%%%%%%%%%%%%%%%%%%%%%%%%%%%%%%%
%%%%%%%%%%%%%%%%%%%%%%%%%%%%%%%%%%%%%%%%%%%%%%%%%%%%%%%%%%%%%%%%%%%%%%%%%%%%%%%
\section{Conclusion}\label{sec:conclusion}

\vspace{-0.7em}

In this work we addressed the challenge of improving traditional, amortized inference in VAEs using a mixture of inference networks approach.  We demonstrated that this method is both effective in increasing the accuracy of inference and computationally efficient, compared to state-of-the-art semi-amortized inference approaches.  This is, in part, due to the effectiveness of the functional recursive mixture learning algorithm we devise and the nature of the inference model, which does not need to be adapted during the test phase.  As a consequence, our approach yields higher test data likelihood than the competing approaches on several benchmark datasets, but remains as computationally efficient as the conventional VAE inference. 
Our recursive model currently requires users to supply the mixture order as an input to the algorithm. In our future work, we aim to investigate principled ways of selecting the mixture order (i.e., model augmentation stopping criteria). 
We also seek to apply our model to domains with structured data, including sequences (e.g., videos, natural language sentences) and graphs (e.g., molecules, 3D shapes). 

%\hl{maybe add some futer work ideas here.}

%%%%%%%%%%%%%%%%%%%%%%%%%%%%%%%%%%%%%%%%%%%%%%%%%%%%%%%%%%%%%%%%%%%%%%%%%%%%%%%
%%%%%%%%%%%%%%%%%%%%%%%%%%%%%%%%%%%%%%%%%%%%%%%%%%%%%%%%%%%%%%%%%%%%%%%%%%%%%%%

\section*{Broader Impact}

\begin{enumerate}
    \item \textbf{Who may benefit from this research?} For any individuals, practitioners, organizations, and groups who aim to identify the underlying generative process of the high-dimensional structured data via the variational auto-encoding model framework, this research can be a very useful tool that provides highly accurate solutions generalizable to unseen data. 
    \item \textbf{Who may be put at disadvantage from this research?} Not particularly applicable. 
    \item \textbf{What are the consequences of failure of the system?} Any failure of the system that implements our algorithm would not do any serious harm since the failure can be easily detectable at the validation stage, in which case alternative strategies or internal decisions might be looked for.  
    \item \textbf{Whether the task/method leverages biases in the data?} Our method does not leverage biases in the data. 
\end{enumerate}

\begin{comment}
Authors are required to include a statement of the broader impact of their work, including its ethical aspects and future societal consequences. 
Authors should discuss both positive and negative outcomes, if any. For instance, authors should discuss a) 
who may benefit from this research, b) who may be put at disadvantage from this research, c) what are the consequences of failure of the system, and d) whether the task/method leverages
biases in the data. If authors believe this is not applicable to them, authors can simply state this.

Use unnumbered first level headings for this section, which should go at the end of the paper. {\bf Note that this section does not count towards the eight pages of content that are allowed.}

\begin{ack}
Use unnumbered first level headings for the acknowledgments. All acknowledgments
go at the end of the paper before the list of references. Moreover, you are required to declare 
funding (financial activities supporting the submitted work) and competing interests (related financial activities outside the submitted work). 
More information about this disclosure can be found at: \url{https://neurips.cc/Conferences/2020/PaperInformation/FundingDisclosure}.

Do {\bf not} include this section in the anonymized submission, only in the final paper. You can use the \texttt{ack} environment provided in the style file to autmoatically hide this section in the anonymized submission.
\end{ack}
\end{comment}

\newpage 

\begin{center}
%\vspace*{\fill}
\LARGE Supplementary Material
%\vspace*{\fill}
\end{center}

This supplement consists of the following materials: 
%%%%
\begin{itemize}
\item Detailed experimental setups (Sec.~\ref{sec:expmt_setup}).
    \begin{itemize}
    \item Summary of competing approaches (Sec.~\ref{sec:competings})
    \item Summary of datasets (Sec.~\ref{sec:datasets})
    \item Network architectures  (Sec.~\ref{sec:architectures}) 
    \item Experimental setups  (Sec.~\ref{sec:expmt_setups})
    \end{itemize}
\item Experimental results (Sec.~\ref{sec:full_results}).
    \begin{itemize}
    \item Test inference time (Sec.~\ref{sec:inf_time})
    \end{itemize}
\item Comparison with fully-connected decoder networks (Sec.~\ref{sec:fullcon_decoder}).
\item Pseudo Codes (Sec.~\ref{sec:code}).
\end{itemize}
%%%%

%%%%%%%%%%%%%%%%%%%%%%%%%%%%%%%%%%%%%%%%%%%%%%%%%%%%%%%%%%%%%%%%%%%%%%%%%%%%%%%
%%%%%%%%%%%%%%%%%%%%%%%%%%%%%%%%%%%%%%%%%%%%%%%%%%%%%%%%%%%%%%%%%%%%%%%%%%%%%%%
\section{Detailed Experimental Setups}\label{sec:expmt_setup}

%%%%%%%%%%%%%%%%%%%%%%%%%%%%%%%%%%%%%%%%%%%%%%%%%%%%%%%%%%%%%%%%%%%%%%%%%%%
\subsection{Competing Approaches}\label{sec:competings}

The competing approaches are summarized as follows:
%%%%
\begin{itemize}
\item \textbf{VAE}: The standard VAE model (amortized inference)~\cite{vae14,vae14r}.
\item \textbf{SA}: The semi-amortized VAE~\cite{savae_ykim}. We fix the SVI gradient step size as $10^{-3}$, but vary the number of SVI steps from $\{1, 2, 4, 8\}$. 
\item \textbf{IAF}: The autoregressive-based flow model for the encoder $q({\bf z}|{\bf x})$~\cite{iafvae}, which has richer expressiveness than VAE's post-Gaussian encoder. % adopted in VAE models. 
The number of flows %(i.e., the number of compositions) 
is chosen from $\{1, 2, 4, 8\}$. 
\item \textbf{HF}: The Householder flow encoder model that represents the full covariance using the Householder transformation~\cite{hfvae}. The number of flows %(i.e., the number of compositions) 
is chosen from $\{1, 2, 4, 8\}$. 
\item \textbf{ME}: For a baseline comparison, % with our recursive mixture estimation, 
we also consider the same mixture encoder model, but unlike our recursive mixture learning, the model is trained conventionally, end-to-end; all mixture components' parameters are updated simultaneously. The number of mixture components is chosen from $\{2,3,4,5\}$. 
\item \textbf{RME}: Our proposed recursive mixture encoder model. We  vary the number of the components to be added $M$ from $\{1,2,3,4\}$, leading to mixture order $2$ to $5$. 
\end{itemize}
%%%%
%
In addition, we test our RME model modified to employ the previous Boosted VI's entropy regularization schemes. More specifically, we replace our bounded KL loss with the two entropy regularization methods as follows:
%
%%%%
\begin{itemize}
\item \textbf{BVI-ER1}: Following~\cite{bvi_locatello_nips}, we replace our bounded KL loss by $\nu \cdot \mathbb{E}_{q({\bf z}|{\bf x})}[-\log q({\bf z}|{\bf x})]$ estimated by Monte Carlo, where $\nu = 1/\sqrt{t+1}$ %\frac{1}{\sqrt{t+1}}$ 
is the impact that decreases as the training iteration $t$.
\item \textbf{BVI-ER2}: Instead of the Monte Carlo estimation of the entropy, we use \cite{bvi_guo}'s closed-form Gaussian entropy $\log\det\bm{\Sigma}$ where $\bm{\Sigma}$ is the (diagonal) covariance of the new component $q({\bf z}|{\bf x})$. 
\end{itemize}
%%%%

%%%%%%%%%%%%%%%%%%%%%%%%%%%%%%%%%%%%%%%%%%%%%%%%%%%%%%%%%%%%%%%%%%%%%%%%%%%
\subsection{Datasets}\label{sec:datasets}

The following benchmark datasets are used. We randomly hold out $10\%$ of the training data as validation sets, except for \textbf{CelebA}. 
%%%%
\begin{itemize}
    \item \textbf{MNIST}~\cite{mnist}: $60,000$ training images and $10,000$ test images where each image is of dimension $(28 \times 28 \times 1)$. 
    \item \textbf{OMNIGLOT}~\cite{omniglot}: $24,345$ training images and $8,070$ test images where each image is of dimension $(28 \times 28 \times 1)$. 
    %\item \textbf{FashionMNIST}~\cite{fashion_mnist}: $60,000$ training images and $10,000$ test images where each image is of dimension $(28 \times 28 \times 1)$. We randomly hold out $10\%$ of the training set as a validation set. 
    \item \textbf{CIFAR10}~\cite{cifar10}: $50,000$ training images and $10,000$ test images where each image is of dimension $(32 \times 32 \times 3)$. 
    \item \textbf{SVHN}~\cite{svhn}: $73,257$ training images and $26,032$ test images where each image is of dimension $(32 \times 32 \times 3)$. 
    \item \textbf{CelebA}~\cite{celeba}: $202,599$ tightly cropped face images of size $(64 \times 64 \times 3)$. We randomly split the data into $80\%/10\%/10\%$ train/validation/test sets. 
\end{itemize}
%%%%

%%%%%%%%%%%%%%%%%%%%%%%%%%%%%%%%%%%%%%%%%%%%%%%%%%%%%%%%%%%%%%%%%%%%%%%%%%%
\subsection{Network Architectures}\label{sec:architectures}

We adopt the convolutional neural networks for both the encoder and decoder %\footnote{More precisely, we used the convolutional network for the encoder and the transposed convolution network for the decoder.} 
models for all competing approaches. This is because the convolutional networks are believed to outperform fully connected networks for many tasks in the image domain~\cite{cnn_imagenet,cnn_szegedy,cnn_dcgan}. 
We also provide empirical evidence in Sec.~\ref{sec:fullcon_decoder} of this Supplement that the fully-connected decoder architecture is inferior to the deconvnet decoder that we adopted, when the two architectures have roughly equal numbers of parameters. This is why we excluded comparison with the recent Laplacian approximation approach of~\cite{vlae} in the main paper. They use the first-order approximate solver method to obtain the mode of the true posterior, but such linearization of a deep network is only computationally feasible for {\em fully connected} decoder models. On the other hand, our recursive mixture learning admits arbitrary types of encoder/decoder architectures, which is another advantage. In Sec.~\ref{sec:fullcon_decoder} of this Supplement we empirically compare the performance between the Laplace approximation~\cite{vlae} and our approach. % with fully connected decoder models.

For the encoder architecture, we first apply $L$ convolutional layers with $(4 \times 4)$-pixels kernels, followed by two fully-connected layers with hidden layers dimension $h$. For the decoder, the input images first go through two fully connected layers, followed by $L$ deconvolution ({\em transposed convolution}) layers with $(4 \times 4)$-pixels filters. Here, $L=3$ for all datasets except CelebA which has $L=4$. The hidden layer dimension $h=256$ for MNIST/OMNIGLOT and $h=512$ for the others. 
%Each component in our recursive mixture encoder is modeled as convolutional networks. 
For fair comparison, the same convolutional network architectures are used in all competing methods. % including VAE, SA (semi-amortized approach), and also the base density in the flow-based models, IAF and HF.

For our recursive mixture RME, all mixture components of the inference model are initialized identically with the VAE's encoder. For the ME (blind end-to-end mixture learning), the first mixture component is initialized with the VAE's encoder while the others are chosen randomly. This is because initializing all components identically would constitute a local maximum of the log-likelihood objective function of the ME, making it unable to update the model further. 
For the IAF, we follow the inverse autoregressive flow modeling~\cite{iafvae} where we use the two-layer MADE~\cite{made} (with the number of hidden units 500) as the autoregressiveNN network. The base density, which is transformed to a more complex density by the flow, is initialized with the trained VAE's encoder $q({\bf z}|{\bf x})$. For the HF, the latents of the base encoder go through a number of linear transformations, followed by the Householder transformation, where the base encoder is also initialized with the VAE's encoder.

The decoder is modeled as transposed convolutional networks. The network architectures are slightly different across the datasets due to different input image dimensions. We summarize the full network architectures in \autoref{tab:arch_1x28x28} (MNIST and OMNIGLOT), \autoref{tab:arch_3x32x32} (CIFAR10 and SVHN), and \autoref{tab:arch_3x64x64} (CelebA). 

%For the mixing proportion network $\epsilon({\bf x};\bm{\eta})$ in our recursive mixture learning method (RME), we used a fully connected network with one hidden layer of dimension $10$. 

In our recursive mixture model, we also need to define the impact function $\epsilon({\bf x})$ for each component. We used a fully connected network $\epsilon({\bf x};\bm{\eta})$ with one hidden layer of dimension $10$. To prevent a new component from overly taking the mixing proportion, we set an upper bound $\epsilon_{\max}$ on the output of the network. This is done by applying the sigmoid function to the output of $\epsilon({\bf x})$, and multiplication by $\epsilon_{\max}$. For all our experiments $\epsilon_{\max}=0.1$ worked well.

%%%%%%%%%%%%%%%%%%%%%%%%%%%%%%%%%%%%%%%%%%%%%%%%%%%%%%%%%%%%%%%%%%%%%%%%%%%
\subsection{Experimental Setups}\label{sec:expmt_setups}

For all optimization, we used the Adam optimizer with batch size $128$ and learning rate $0.0005$. We run the optimization until 2000 epochs. 
We vary the latent dimension $\textrm{dim}({\bf z})$, %In the previous work~\cite{cremer18,vlae}, they used fully connected networks for encoder/decoder with two different complexities, small and large. But it turns out that convnet significantly outperforms the full-con nets provided that the number of parameters of the two architectures are (roughly) equal. Instead, what matters more is the latent dimension, hence we vary the $\textrm{dim}({\bf z})$ from small to large. 
from $\{10,20,50,100\}$. To report the test log-likelihood scores $\log p({\bf x})$, we use the importance weighted sampling estimation (IWAE) method~\cite{iwae}. More specifically, 
%%%%
\begin{equation}
\textrm{IWAE} = \log \Bigg( \frac{1}{K} \sum_{i=1}^K 
  \frac{ p({\bf x},{\bf z}_i) } { q({\bf z}_i|{\bf x}) } \Bigg),
\label{eq:iwae}
\end{equation}
%%%%
where ${\bf z}_1,\dots,{\bf z}_K$ are i.i.d.~samples from $q({\bf z}|{\bf x})$. It can be shown that IWAE lower bounds $\log p({\bf x})$ and can be arbitrarily close to the target as the number of samples $K$ grows. We use $K=100$ throughout the experiments. 

For each model/dataset, we perform 10 runs with different random train/validation splits, where each run consists of three trainings by starting with different random model parameters, among which only one model with the highest validation performance is chosen. To see the statistical significance of difference between competing models, we also performed the one-sided Wilcoxon signed rank test for every pair, namely (the best model vs.~each non-best model), using the 10 log-likelihood scores per model. 

%%%%%%%%
\begin{table*}%[h!]
\caption{Encoder (i.e., each component in our mixture model) and decoder network architectures for MNIST and OMNIGLOT datasets. In the convolutional and transposed convolutional layers, the paddings are properly adjusted to match the input/output dimensions.}
\label{tab:arch_1x28x28}
\vspace{-1.0em}
\begin{center}
\begin{small}
\begin{sc}
\begin{tabular}{l|l}
\toprule
 Encoder & Decoder \\
\midrule
Input: $(28 \times 28 \times 1)$ & Input: ${\bf z} \in $ $\mathbb{R}^{p}$ ($p\in\{10,20,50,100\})$ \\
\midrule
32 (4 $\times$ 4) conv.; stride 2; LeakyReLU ($0.01$) & FC. 256; ReLU \\
\midrule
32 (4 $\times$ 4) conv.; stride 2; LeakyReLU ($0.01$) & FC. $3 \cdot 3 \cdot 64$; RELU \\
\midrule
64 (4 $\times$ 4) conv.; stride 2; LeakyReLU ($0.01$) & 32 (4 $\times$ 4) Transposed Conv.; stride 2; ReLU \\
\midrule
FC. 256; LeakyReLU ($0.01$) & 32 (4 $\times$ 4) Transposed Conv.; stride 2; ReLU \\
\midrule
FC. 2 $\times p$ ($p=\textrm{dim}({\bf z})\in\{10,20,50,100\})$ & 1 (4 $\times$ 4) Transposed Conv.; stride 2 \\
\bottomrule
\end{tabular}
\end{sc}
\end{small}
\end{center}
\vskip -0.1in
%\vspace{-1.0em}
\end{table*}
%%%%%%%%
%
%%%%%%%%
\begin{table*}%[h!]
\caption{Encoder %(i.e., each component in our mixture model)
and decoder network architectures for CIFAR10 and SVHN datasets. %In the convolutional and transposed convolutional layers, the paddings are properly adjusted to match the input/output dimensions.
}
\label{tab:arch_3x32x32}
\vspace{-1.0em}
\begin{center}
\begin{small}
\begin{sc}
\begin{tabular}{l|l}
\toprule
 Encoder & Decoder \\
\midrule
Input: $(32 \times 32 \times 3)$ & Input: ${\bf z} \in $ $\mathbb{R}^{p}$ ($p\in\{10,20,50,100\})$ \\
\midrule
32 (4 $\times$ 4) conv.; stride 2; LeakyReLU ($0.01$) & 
FC. 512; ReLU \\
\midrule
32 (4 $\times$ 4) conv.; stride 2; LeakyReLU ($0.01$) & 
FC. $4 \cdot 4 \cdot 64$; RELU \\
\midrule
64 (4 $\times$ 4) conv.; stride 2; LeakyReLU ($0.01$) & 
32 (4 $\times$ 4) Transposed Conv.; stride 2; ReLU \\
\midrule
FC. 512; LeakyReLU ($0.01$) & 
32 (4 $\times$ 4) Transposed Conv.; stride 2; ReLU \\
\midrule
FC. 2 $\times p$ ($p=\textrm{dim}({\bf z})\in\{10,20,50,100\})$ & 
3 (4 $\times$ 4) Transposed Conv.; stride 2 \\
\bottomrule
\end{tabular}
\end{sc}
\end{small}
\end{center}
\vskip -0.1in
%\vspace{-1.0em}
\end{table*}
%%%%%%%%
%
%%%%%%%%
\begin{table*}%[h!]
\caption{Encoder %(i.e., each component in our mixture model)
and decoder network architectures for CelebA dataset. %In the convolutional and transposed convolutional layers, the paddings are properly adjusted to match the input/output dimensions.
}
\label{tab:arch_3x64x64}
\vspace{-1.0em}
\begin{center}
\begin{small}
\begin{sc}
\begin{tabular}{l|l}
\toprule
 Encoder & Decoder \\
\midrule
Input: $(64 \times 64 \times 3)$ & Input: ${\bf z} \in $ $\mathbb{R}^{p}$ ($p\in\{10,20,50,100\})$ \\
\midrule
32 (4 $\times$ 4) conv.; stride 2; LeakyReLU ($0.01$) & 
FC. 512; ReLU \\
\midrule
32 (4 $\times$ 4) conv.; stride 2; LeakyReLU ($0.01$) & 
FC. $4 \cdot 4 \cdot 64$; RELU \\
\midrule
64 (4 $\times$ 4) conv.; stride 2; LeakyReLU ($0.01$) & 
64 (4 $\times$ 4) Transposed Conv.; stride 2; ReLU \\
\midrule
64 (4 $\times$ 4) conv.; stride 2; LeakyReLU ($0.01$) & 
32 (4 $\times$ 4) Transposed Conv.; stride 2; ReLU \\
\midrule
FC. 512; LeakyReLU ($0.01$) & 
32 (4 $\times$ 4) Transposed Conv.; stride 2; ReLU \\
\midrule
FC. 2 $\times p$ ($p=\textrm{dim}({\bf z})\in\{10,20,50,100\})$ & 
3 (4 $\times$ 4) Transposed Conv.; stride 2 \\
\bottomrule
\end{tabular}
\end{sc}
\end{small}
\end{center}
%\vskip -0.1in
%\vspace{-1.0em}
\end{table*}
%%%%%%%%

%%%%%%%%%%%%%%%%%%%%%%%%%%%%%%%%%%%%%%%%%%%%%%%%%%%%%%%%%%%%%%%%%%%%%%%%%%%%%%%
%%%%%%%%%%%%%%%%%%%%%%%%%%%%%%%%%%%%%%%%%%%%%%%%%%%%%%%%%%%%%%%%%%%%%%%%%%%%%%%
\section{Experimental Results}\label{sec:full_results}

The test log-likelihood scores are summarized in \autoref{tab:mnist} (MNIST)\footnote{For the MNIST results, the test log-likelihood scores of the competing methods mismatch those reported in the related work (e.g.,~\cite{vampprior}). Significantly higher scores. This is because we adopt the Gaussian decoder models, not the binary decoders, for all competing methods.}, \autoref{tab:omniglot} (OMNIGLOT), \autoref{tab:cifar10} (CIFAR10), \autoref{tab:svhn} (SVHN), and \autoref{tab:celeba} (CelebA). %The qualitative results of reconstructed and synthesized images can be found in the Supplement. 
% Analysis/interpretation for each dataset... 
We also report the performance of the 
entropy regularization schemes introduced in the previous Boosted VI (BVI) approaches. To this end, in our RME, we replace our bounded KL (BKL) loss with the entropy regularization. More specifically, we consider two entropy regularization schemes -- \textbf{BVI-ER1}: \cite{bvi_locatello_nips}'s regularization of the negative entropy of $q({\bf z}|{\bf x})$ whose impact decreases $\frac{1}{\sqrt{t+1}}$ as a function of training iteration $t$, as suggested. 
\textbf{BVI-ER2}: \cite{bvi_guo}'s Gaussian entropy based regularization (i.e., penalizing small $\log\det\bm{\Sigma}$ where $\bm{\Sigma}$ is the (diagonal) covariance matrix of the new component $q({\bf z}|{\bf x})$ to be optimized. 
Overall the results indicate that our recursive mixture encoder (RME) outperforms the competing approaches consistently for all datasets.

\subsection{Test Inference Time}\label{sec:inf_time}

Another key advantage of our recursive mixture model is the computational efficiency of test-time inference, comparable to that of VAE. Unlike the semi-amortized approaches, where one performs the SVI gradient adaptation at test time, the inference in our RME is merely a single feed forward pass through our mixture encoder network. That is, once training is done, our mixture inference model remains fixed, with no adaptation required.

%%%%
\begin{table}[t!]
\centering
\caption{(Per-batch) Test inference time (in milliseconds) with batch size 128. The latent dimension $\textrm{dim}({\bf z})=50$. %MNIST & OMNIGLOT & CIFAR10 & SVHN & CelebA
}
\label{tab:inf_time}
%\vskip 0.05in
\begin{small}
\begin{sc}
\centering
\begin{tabular}{lccccc}
\toprule
%& M & O & Ci & S & Ce \\
& {MNIST} & {OMNIG.} & {CIFAR10} & {SVHN} & {CelebA} \\
\midrule
VAE & \ \ 3.6 & \ \ 4.8 & \ \ 3.7 & \ \ 2.2 & \ \ 2.7 \\ \hline
SA ($1$) & \ \ 9.7 & 11.6 & \ \ 9.8 & \ \ 7.0 & \ \ 8.4 \\
SA ($2$) & 18.1 & 19.2 & 16.8 & 15.5 & 13.8 \\
SA ($4$) & 32.2 & 34.4 & 27.9 & 30.1 & 27.1 \\
SA ($8$) & 60.8 & 65.7 & 60.5 & 60.3 & 53.8 \\ \hline
IAF ($1$) & \ \ 4.8 & \ \ 5.7 & \ \ 5.1 & \ \ 3.4 & \ \ 4.4 \\
IAF ($2$) & \ \ 5.9 & \ \ 6.4 & \ \ 5.6 & \ \ 3.7 & \ \ 5.1 \\
IAF ($4$) & \ \ 6.2 & \ \ 7.0 & \ \ 6.3 & \ \ 4.7 & \ \ 5.7 \\
IAF ($8$) & \ \ 7.7 & \ \ 8.2 & \ \ 7.6 & \ \ 5.7 & \ \ 7.7 \\ \hline
RME ($2$) & \ \ 4.7 & \ \ 5.4 & \ \ 4.9 & \ \ 3.2 & \ \ 4.2 \\
RME ($3$) & \ \ 4.9 & \ \ 5.5 & \ \ 5.1 & \ \ 3.6 & \ \ 4.1 \\
RME ($4$) & \ \ 4.6 & \ \ 5.3 & \ \ 5.1 & \ \ 3.5 & \ \ 4.2 \\
RME ($5$) & \ \ 4.8 & \ \ 5.6 & \ \ 5.1 & \ \ 3.3 & \ \ 4.8 \\ 
\bottomrule
\end{tabular}
\end{sc}
\end{small}
\end{table}
%%%%

To verify this, we measure the actual inference time for competing approaches. The per-batch inference times (batch size 128) on all benchmark datasets are shown in \autoref{tab:inf_time}. To report the results, for each method and each dataset, we run the inference over the entire test set batches, measure the running time, then take the per-batch average. We repeat the procedure five times and report the average. All models are run on the same machine with a single GPU (RTX 2080 Ti), Core i7 3.50GHz CPU, and 
128 GB RAM. We only report test times for the latent dimension $\textrm{dim}({\bf z})=50$ %(the highest dimension in our experimental setup) 
as the impact of the latent dimension appears to be less significant. 

As expected, the semi-amortized approach (SA) suffers from the computational overhead of test time gradient updates, with the inference time significantly increased as the number of the updates increases. Our RME is comparable to the VAE, and faster than the IAF (with more than a single flow), which verifies our claim. Interestingly, increasing the mixture order in our model rarely affects the inference time, due to intrinsic parallelization of the feed forward pass through the multiple mixture components networks, leading to inference times as fast as those of the single component model (VAE). %, where the number of mixture components do not really matter. 

%%%%%%%%%%%%%%%%%%%%%%%%%%%%%%%%%%%%%%%%%%%%%%%%%%%%%%%%%%%%%%%%%%%%%%%%%%%%%%%
%%%%%%%%%%%%%%%%%%%%%%%%%%%%%%%%%%%%%%%%%%%%%%%%%%%%%%%%%%%%%%%%%%%%%%%%%%%%%%%
\section{Comparison with Fully-Connected Decoder Networks}\label{sec:fullcon_decoder}

In the main paper we used the convolutional networks for both encoder and decoder models. This is a reasonable architectural choice considering that all the datasets are images. Also it is widely believed that convolutional networks outperform fully connected networks for many tasks in the image domain~\cite{cnn_imagenet,cnn_szegedy,cnn_dcgan}. However, one can alternatively consider fully connected networks for either the encoder or the decoder, or both. Nevertheless, being equal in the number of model parameters, using both convolutional encoder and decoder networks always outperformed the fully connected counterparts. %More specifically, the model equipped with fully connected encoder and decoder networks is the worst, then the model having convolutional network for either encoder of decoder is better, and using convnets for both encoder and decoder performs the best.  
In this section we empirically verify this by comparing the test likelihood performance between the two architectures. We particularly focus on comparing the two architectures (convolutional vs.~fully connected) for the {\em decoder} model alone, while retaining the convolutional network encoder for both cases. 

Using the fully connected decoder network allows us to test the recent Laplacian approximation approach~\cite{vlae} (denoted by \textbf{VLAE}), which we excluded from the main paper. They employ a first-order approximation solver to find the mode of the true posterior (i.e., linearizing the decoder function), and compute the Hessian of the log-posterior at the mode to define the  (full) covariance matrix. This procedure is  computationally feasible only for a fully connected decoder model. We conduct experiments on MNIST and OMNIGLOT datasets where the fully connected decoder network consists of two hidden layers and the hidden layer dimensions are chosen to set the total number of weight parameters roughly equal to the convolutional decoder network used in the main paper. 

\autoref{tab:fullcon_decoder} summarizes the results. Among the fully connected networks, the VLAE achieves the highest performance. Instead of doing SVI gradient updates as in the SAVI method (SA), the VLAE aims to directly solve for the mode of the true posterior by decoder linearization, leading to more accurate posterior refinement without suffering from the step size issue. % (and the Hessian of the true negative log posterior at the mode point), and set them as the mean and covariance of the encoder. The mode seeking is done by the first-order approximate solver method, which somehow circumvents the SAVI's step size issue. 
Our recursive mixture, with the fully connected decoder networks, still improves the VAE's scores, but the improvement is often less than that of the VLAE. However, when compared to the convnet decoder cases, even the conventional VAE significantly outperforms the VLAE. The best VLAE's scores are significantly lower than VAE's using convolutional decoders. Restricted network architecture of the VLAE is its main drawback. 

%%%%
\begin{table}[t!]
\centering
\caption{%(OMNIGLOT with 
(Fully connected vs.~convolutional decoder networks) Test log-likelihood scores (unit in nat). The figures without parentheses are the scores using the fully connected networks, whereas figures in the parentheses are the scores using the convolutional decoder networks. Both architectures have roughly equal number of the weight parameters. The number of linearization steps in the VLAE is chosen from $\{1,2,4,8\}$.
}
\label{tab:fullcon_decoder}
\vspace{-0.5em}
%\vskip 0.05in
\begin{small}
\begin{sc}
\centering
\begin{tabular}{l|cc|cc}
\toprule
 & \multicolumn{2}{c}{MNIST} & \multicolumn{2}{|c}{OMNIGLOT} \\
 \cline{2-5}
 & \ \ $\textrm{dim}({\bf z})=10$ \ \ & \ \ $\textrm{dim}({\bf z})=50$ \ \ & \ \ $\textrm{dim}({\bf z})=10$ \ \ & \ \ $\textrm{dim}({\bf z})=50$ \ \ \\
\midrule
VAE & 563.6 (685.1) & 872.6 (1185.7) & 296.8 (347.0) & 519.4 (801.6) \\ \hline
SA ($1$) & 565.1 (688.1) & 865.8 (1172.1) & 297.6 (344.1) & 489.0 (792.7) \\
SA ($2$) & 565.3 (682.2) & 868.2 (1176.3) & 295.3 (349.5) & 534.1 (793.1) \\
SA ($4$) & 565.9 (683.5) & 852.9 (1171.3) & 294.8 (342.1) & 497.8 (794.4) \\
SA ($8$) & 564.9 (684.6) & 870.9 (1183.2) & 299.0 (344.8) & 500.0 (799.4) \\ \hline
VLAE ($1$) & 590.0 & 922.2 & 307.4 & 644.0 \\
VLAE ($2$) & 595.1 & 908.8 & 307.6 & 621.4 \\
VLAE ($4$) & 605.2 & 841.4 & 318.0 & 597.7 \\
VLAE ($8$) & 605.7 & 779.9 & 316.6 & 553.1  \\ \hline
RME ($2$) & 570.9 (697.2) & 888.1 (1201.7) & 298.4 (349.3) & 524.7 (821.0) \\
RME ($3$) & 571.9 (698.2) & 888.2 (1202.4) & 298.6 (349.9) & 524.8 (820.4) \\
RME ($4$) & 571.4 (699.0) & 888.1 (1203.1) & 298.8 (350.7) & 525.3 (819.9) \\
RME ($5$) & 572.2 (699.4) & 888.0 (1203.7) & 298.8 (351.1) & 526.9 (819.9) \\
\bottomrule
\end{tabular}
\end{sc}
\end{small}
\end{table}
%%%%

We also compare the test inference times of our recursive mixture model and the VLAE using the fully connected decoder networks. Note that VLAE is a semi-amortized approach, which needs to solve the Laplace approximation at test time. Thus another drawback of VLAE is the computational overhead of inference,  which can be demanding as the number of linearization steps increases. The per-batch inference times (batch size 128) are shown in \autoref{tab:time_fullcon_decoder}. For the moderate or large linearization steps (e.g., 4 or 8), the inference takes significantly longer than that of our RME (amortized method). 

%%%%
\begin{table}[t]
\centering
\caption{%(OMNIGLOT with 
(Fully connected networks as decoders) 
Per-batch inference time (unit in milliseconds) with batch size 128. %For the VLAE, we use the fully connected decoder with the same number of parameters as the convnet decoder. 
The figures without parentheses are the times using the fully connected networks, whereas figures in the parentheses are the times using the convolutional decoder networks. %Both architectures have roughly equal number of weight parameters.
}
\label{tab:time_fullcon_decoder}
\vspace{-0.5em}
%\vskip 0.05in
\begin{small}
\begin{sc}
\centering
\begin{tabular}{l|cc|cc}
\toprule
 & \multicolumn{2}{c}{MNIST} & \multicolumn{2}{|c}{OMNIGLOT} \\
 \cline{2-5}
 & \ \ $\textrm{dim}({\bf z})=10$ \ \ & \ \ $\textrm{dim}({\bf z})=50$ \ \ & \ \ $\textrm{dim}({\bf z})=10$ \ \ & \ \ $\textrm{dim}({\bf z})=50$ \ \ \\
\midrule
VLAE ($1$) & 10.1 & 12.9 & 11.2 & 12.1 \\
VLAE ($2$) & 11.2 & 13.4 & 13.2 & 16.9 \\
VLAE ($4$) & 14.8 & 17.8 & 15.4 & 18.7 \\
VLAE ($8$) & 20.7 & 30.8 & 22.1 & 26.4 \\ \hline
RME ($2$) & 5.0 (5.0) & 5.0 (4.7) & 5.4 (6.0) & 5.6 (5.4) \\
RME ($3$) & 4.9 (5.1) & 4.9 (4.9) & 5.9 (5.7) & 5.4 (5.5) \\
RME ($4$) & 4.9 (5.0) & 4.9 (4.6) & 6.1 (5.9) & 5.9 (5.3) \\
RME ($5$) & 5.0 (5.1) & 4.7 (4.8) & 5.8 (6.1) & 5.4 (5.6) \\
\bottomrule
\end{tabular}
\end{sc}
\end{small}
\end{table}
%%%%

%%%%%%%%%%%%%%%%%%%%%%%%%%%%%%%%%%%%%%%%%%%%%%%%%%%%%%%%%%%%%%%%%%%%%%%%%%%%%%%
%%%%%%%%%%%%%%%%%%%%%%%%%%%%%%%%%%%%%%%%%%%%%%%%%%%%%%%%%%%%%%%%%%%%%%%%%%%%%%%
\section{Pseudo Codes}\label{sec:code}

The following is the pseudocode for the proposed model. % based in Python/PyTorch. %A text file version of this pseudocode is also attached in the supplement. 
The real full Python/PyTorch code %can be publicly 
is available 
in \url{https://github.com/minyoungkim21/recmixvae}. 
%should the paper be accepted. 

%\vspace{+2.0em}

\begin{footnotesize}
\begin{verbatim}
  #### Hyperparameters ####

  batch_size = 128                # input batch size for training
  n_epochs = 2000                 # number of epochs to train
  x_dim = (C=1 x H=28 x W=28)     # input dimension
  z_dim = 50                      # latent space dimension
  learning_rate = 1e-6            # learning rate for ADAM optimizer

  num_comps = 5                   # number of mixture components for encoder
  eps_regr_nhl = 1                # number of hidden layers for epsilon regressor
  eps_regr_dim = 10               # hidden layer dim for epsilon regressor
  eps_min = 0.001                 # minimum epsilon 
  eps_max = 0.1                   # maximum epsilon
  kl_max = 500.0                  # maximum kl(q_k||Q_{k-1}) allowed in the objective


  #### Main class ####

  import torch.nn as nn

  class RecMixVAE(nn.Module):
  
      self.M = num_comps-1  # components: 0,1,...,M (the number of comps = M+1)
      self.decoder = ConvDecoder(z_dim, x_dim)  # decoder
      self.prior = DiagonalGaussian(mu=zeros, logvar=zeros)  # prior
        
      # components of encoder (q_0, q_1, ..., q_M)
      self.comps = nn.ModuleList( [ConvEncoder(z_dim, x_dim) for _ in range(num_comps)] )
      
      # regressors for impacts of components  (eps_0, eps_1, ..., eps_M); note: eps_0 = 1 (const)
      self.eps_regrs = nn.ModuleList( [Const(1.0)] + 
          [ BaseBoundedRegressor( x_dim, eps_min, eps_max, eps_regr_nhl, eps_regr_dim ) 
            for _ in range(num_comps-1) ] )
        
      def encoder_upto_kth(self, x, k):
          '''
          Mixture with components q_0(.|x), q_1(.|x), ..., q_k(.|x) is formed.
          More specifically, eg, for k=2, 
            Q_{k=2}(.|x) = alpha_0(x) * q_0(.|x) + alpha_1(x) * q_1(.|x) + alpha_2(x) * q_2(.|x) 
          where 
            alpha_2(x) = eps_2(x)
            alpha_1(x) = eps_1(x) * (1-eps_2(x))
            alpha_0(x) = eps_0(x) * (1-eps_1(x)) * (1-eps_2(x))
          inputs:
            k = component index (0 <= k <= self.M)
          returns:
            n mixtures for Q_k(.|x) (with k+1 components)
          '''

      def encoder_kth_comp(self, x, k):
          '''
          Just return k-th component q_k(.|x)
          inputs:
            k = component index (0 <= k <= self.M)
          returns:
            n distributions (eg, DiagonalGaussian's) q_k(.|x)
          '''
          return self.comps[k](x)[0]
    
      def eval_elbo_for_mixture(self, x, mixture):
          '''
          Evaluate elbo (recon error and kl) for a mixture encoder
          inputs:
            mixture = n mixture distributions from Q(.|x)
          returns:
            ell = E_{Q(z|x)}[ log p(x|z) ] 
            kl = KL( Q(z|x) || p(z) )
          '''
          let K = mixture order
          alphas = mixture.logalphas.exp()
          z = samples from q_m(z|x) for m=1...K
          (decoder) evaluate log p(x|z) for z ~ q_m(z|x) for m=1...K
          (prior) evaluate log p(z) for z ~ q_m(z|x) for m=1...K
          evaluate log Q(z|x) for z ~ q_m(z|x) for m=1...K
          return ell = E_{Q(z|x)}[ log p(x|z) ] and kl = KL( Q(z|x) || p(z) )
          
      def forward(self, x, k, loss_type):
          '''
          compute objectives for recursive mixture VAE
          inputs:
            k = component index (0 <= k <= self.M)
            loss_type = either of
                'new_comp': compute elbo(q_k) and kl(q_k||Q_{k-1}) (the latter None if k=0)
                'mixture': compute elbo(Q_k)
          returns:
            loss_type == 'new_comp': elbo(q_k), kl(q_k||Q_{k-1}) (averaged over batch x)
            loss_type == 'mixture': elbo(Q_k) (averaged over batch x)
          '''
        if loss_type == 'new_comp':
            q_z_x = self.encoder_kth_comp(x, k)  # q_k
            Q_z_x = self.encoder_upto_kth(x, k-1) if k>0 else None  # Q_{k-1}
            evaluate elbo(q_k) and kl(q_k||Q_{k-1})
        elif loss_type == 'mixture':
            Q_z_x = self.encoder_upto_kth(x, k)  # Q_k
            ell, kl = self.eval_elbo_for_mixture(x, Q_z_x)
            elbo = ( ell - kl ).mean()
            
      def enable_grad(self, params):
          '''
          Disable the autograd for all parameters except for "params"
          '''
            
            
  #### Main algorithm ####
  
  model = RecMixVAE()
  
  while epoch <= n_epochs:
  
      for batch sampled from the training data:
            
          # update q_0
          model.enable_grad(model.comps[0])
          elbo, _ = model(batch, 0, loss_type='new_comp')
          update model by backprop with loss = -elbo
            
          # update (q_m, eps_regr_m) for m=1,...,M
          for m in range(1,model.M+1):
        
              # update q_m
              model.enable_grad(model.comps[m])
              elbo, kl = model(batch, m, loss_type='new_comp')
              update model by backprop with loss = -elbo + (kl_max - kl).relu()
                
              # update eps_regr_m
              model.enable_grad(model.eps_regrs[m])
              elbo = model(batch, m, loss_type='mixture')
              update model by backprop with loss = -elbo
            
          # update decoder
          model.enable_grad(model.decoder)
          elbo = model(batch, model.M, loss_type='mixture')
          update model by backprop with loss = -elbo
        

\end{verbatim}
\end{footnotesize}

%%%%%%%%%%%%%%%
%\subsubsection*{Acknowledgments}

%Use unnumbered third level headings for the acknowledgments. 

% \definecolor{lor}{rgb}{1,0.85,0}
% \definecolor{or}{rgb}{1,0.60,0}
% \definecolor{dor}{rgb}{1,0.20,0}
% \newcommand\Tstrut{\rule{0pt}{2.2ex}}         % = `top' strut
% \newcommand\Bstrut{\rule[-0.9ex]{0pt}{0pt}}   % = `bottom' strut

%%%% 
\begin{table}%[t]
\centering
\caption{(MNIST) Test log-likelihood scores (unit in nat) estimated by the importance weighted sampling~\cite{iwae}. % with 100 samples. 
The figures in the parentheses next to model names indicate: the number of SVI steps in SA, the number of flows in IAF and HF, and the number of mixture components in ME and RME. The superscripts are the standard deviations. 
The best (on average) results are boldfaced in $\color{red} \textrm{\textbf{red}}$. 
In each column, the statistical significance of the difference between the best model (red) and each competing model, is depicted as color: anything non-colored indicates $p\leq 0.01$ %\textrm{black}$ 
(strongly distinguished), 
$p \in (0.01,0.05]$ %$0.01<p\leq 0.05 = 
as $\color{lor} \textrm{yellow-orange}$, 
$p \in (0.05,0.1]$ %$0.05<p\leq 0.10 = 
as $\color{or} \textrm{orange}$, 
$p>0.1$ as $\color{dor} \textrm{red orange}$ (little evidence of difference) by the  Wilcoxon signed rank test. Best viewed in color.
%
%\textbf{Comparison to previous Boosted VI methods}. 
% \textbf{BVI}$^\dagger$: \cite{bvi_guo}'s Gaussian entropy based regularization (i.e., penalizing small $\log\det\bm{\Sigma}$ where $\bm{\Sigma}$ is the (diagonal) covariance matrix of the new component $q({\bf z}|{\bf x})$ to be optimized. 
% \textbf{BVI}$^\diamond$: \cite{bvi_locatello_nips}'s regularization of the negative entropy of $q({\bf z}|{\bf x})$ whose impact decreases $\frac{1}{\sqrt{t+1}}$ as a function of training iteration $t$, as suggested.
}
\label{tab:mnist}
\vskip 0.05in
%\begin{scriptsize}
\begin{small}
%\begin{sc}
\centering
\begin{tabular}{lllll}
\toprule
$\textrm{dim}({\bf z})$ & \multicolumn{1}{c}{$10$} %$\textrm{dim}({\bf z})=10$ 
 & \multicolumn{1}{c}{$20$} & \multicolumn{1}{c}{$50$} & \multicolumn{1}{c}{$100$} \\
 %$\textrm{dim}({\bf z})=50$ & $\textrm{dim}({\bf z})=100$ \\
%\midrule
\hline\hline
VAE\Tstrut & $685.1^{1.8}$ & $930.7^{3.9}$ & $1185.7^{3.9}$ & $1225.4^{4.2}$ \\ \hline
SA$^{(1)}$\Tstrut & $688.1^{2.7}$ & $921.2^{2.3}$ & $1172.1^{1.8}$ & $1196.9^{3.3}$ \\
SA$^{(2)}$ & $682.2^{1.5}$ & $932.0^{2.4}$ & $1176.3^{3.4}$ & $1216.7^{2.9}$ \\
SA$^{(4)}$ & $683.5^{1.5}$ & $925.5^{2.6}$ & $1171.3^{3.5}$ & $1217.7^{3.9}$ \\
SA$^{(8)}$ & $684.6^{1.5}$ & $928.1^{3.9}$ & $1183.2^{3.4}$ & $1211.7^{2.9}$ \\ \hline
IAF$^{(1)}$\Tstrut & $687.3^{1.1}$ & $934.0^{3.3}$ & $1180.6^{2.7}$ & $1213.4^{5.6}$ \\
IAF$^{(2)}$ & $677.7^{1.6}$ & $931.4^{3.7}$ & $1190.1^{1.9}$ & $1224.4^{2.2}$ \\
IAF$^{(4)}$ & $685.0^{1.5}$ & $926.3^{2.6}$ & $1178.1^{1.6}$ & $1216.4^{3.9}$ \\
IAF$^{(8)}$ & $689.7^{1.4}$ & $934.1^{2.4}$ & $1150.0^{2.2}$ & $1190.9^{3.9}$ \\ \hline
HF$^{(1)}$\Tstrut & $682.5^{1.4}$ & $917.2^{2.6}$ & $\color{dor} 1204.3^{4.0}$ & $1203.3^{2.3}$ \\
HF$^{(2)}$ & $677.6^{2.2}$ & $923.9^{3.1}$ & $1191.5^{10.8}$ & $1213.6^{3.0}$ \\
HF$^{(4)}$ & $683.3^{2.6}$ & $927.3^{2.8}$ & $1197.2^{1.5}$ & $1226.0^{2.0}$ \\
HF$^{(8)}$ & $679.6^{1.5}$ & $928.5^{3.1}$ & $1184.1^{1.8}$ & $1220.0^{3.5}$ \\ \hline
ME$^{(2)}$\Tstrut & $685.7^{1.2}$ & $926.7^{3.0}$ & $1152.8^{1.7}$ & $1191.4^{2.5}$ \\
ME$^{(3)}$ & $678.5^{2.5}$ & $933.1^{4.1}$ & $1162.8^{4.7}$ & $1216.9^{2.1}$ \\
ME$^{(4)}$ & $680.0^{0.9}$ & $914.7^{2.3}$ & $\color{red} {\bf 1205.1}^{2.3}$ & $1214.9^{3.4}$ \\
ME$^{(5)}$ & $682.0^{1.7}$ & $920.6^{1.9}$ & $1198.5^{3.5}$ & $1181.7^{3.7}$ \\ \hline
RME$^{(2)}$\Tstrut & $\color{dor} 697.2^{1.1}$ & $\color{lor} 943.9^{1.6}$ & $\color{or} 1201.7^{0.9}$ & $\color{lor} 1240.7^{2.5}$ \\
RME$^{(3)}$ & $\color{dor} 698.2^{1.1}$ & $\color{dor} 945.1^{1.6}$ & $\color{or} 1202.4^{1.0}$ & $\color{lor} 1240.8^{2.4}$ \\
RME$^{(4)}$ & $\color{dor} 699.0^{1.0}$ & $\color{red} {\bf 945.2}^{1.6}$ & $\color{dor} 1203.1^{1.0}$ & $\color{or} 1241.5^{2.4}$ \\
RME$^{(5)}$ & $\color{red} {\bf 699.4}^{2.1}$ & $\color{dor} 945.0^{1.7}$ & $\color{dor} 1203.7^{1.0}$ & $\color{red} {\bf 1242.0}^{2.4}$ \\
\hline\hline
BVI-ER1$^{(2)}$\Tstrut & $694.5^{1.9}$ & $939.7^{2.8}$ & $1196.2^{2.8}$ & $1236.3^{3.0}$ \\
BVI-ER1$^{(3)}$ & $694.5^{1.9}$ & $939.5^{2.9}$ & $1191.6^{2.9}$ & $1233.9^{3.0}$ \\
BVI-ER1$^{(4)}$ & $692.2^{1.8}$ & $937.8^{2.9}$ & $1191.6^{2.8}$ & $1227.6^{3.0}$ \\
BVI-ER1$^{(5)}$ & $692.0^{1.9}$ & $931.2^{3.0}$ & $1183.1^{2.9}$ & $1229.0^{3.1}$ \\
\hline
BVI-ER2$^{(2)}$\Tstrut & $694.5^{1.9}$ & $939.7^{2.1}$ & $1189.6^{2.2}$ & $1236.2^{3.0}$ \\
BVI-ER2$^{(3)}$ & $694.5^{1.9}$ & $939.4^{2.1}$ & $1192.1^{2.3}$ & $1233.6^{3.0}$ \\
BVI-ER2$^{(4)}$ & $692.2^{1.9}$ & $937.6^{2.1}$ & $1191.5^{2.2}$ & $1227.4^{3.0}$ \\
BVI-ER2$^{(5)}$ & $692.4^{1.9}$ & $931.7^{2.2}$ & $1181.7^{2.2}$ & $1228.9^{3.0}$ \\
% \hline
%
\bottomrule
\end{tabular}
%\end{sc}
%\end{scriptsize}
\end{small}
\end{table}
%%%%

%%%%
\begin{table}%[b]
\centering
\caption{(OMNIGLOT) Test log-likelihood scores (unit in nat). The same interpretation as \autoref{tab:mnist}. 
}
\label{tab:omniglot}
\vskip 0.05in
\begin{small}
%\begin{sc}
\centering
\begin{tabular}{lllll}
\toprule
$\textrm{dim}({\bf z})$ & \multicolumn{1}{c}{$10$} %$\textrm{dim}({\bf z})=10$ 
 & \multicolumn{1}{c}{$20$} & \multicolumn{1}{c}{$50$} & \multicolumn{1}{c}{$100$} \\
 %$\textrm{dim}({\bf z})=50$ & $\textrm{dim}({\bf z})=100$ \\
%\midrule
\hline\hline
VAE\Tstrut & $347.0^{1.7}$ & $501.6^{1.6}$ & $801.6^{4.0}$ & $917.5^{5.1}$ \\ \hline
SA$^{(1)}$\Tstrut & $344.1^{1.4}$ & $499.3^{2.5}$ & $792.7^{7.9}$ & $905.8^{4.2}$ \\
SA$^{(2)}$ & $\color{dor} 349.5^{1.4}$ & $501.0^{2.7}$ & $793.1^{4.8}$ & $920.0^{4.5}$ \\
SA$^{(4)}$ & $342.1^{1.0}$ & $488.2^{1.8}$ & $794.4^{1.9}$ & $914.6^{5.6}$ \\
SA$^{(8)}$ & $344.8^{1.1}$ & $490.3^{2.8}$ & $799.4^{2.7}$ & $\color{lor} 942.2^{5.2}$ \\ \hline
IAF$^{(1)}$\Tstrut & $\color{lor} 347.8^{1.6}$ & $489.9^{1.9}$ & $788.8^{4.1}$ & $937.4^{7.2}$ \\
IAF$^{(2)}$ & $344.2^{1.6}$ & $494.9^{1.4}$ & $795.7^{2.7}$ & $934.6^{7.3}$ \\
IAF$^{(4)}$ & $347.9^{1.9}$ & $496.0^{2.0}$ & $775.1^{2.2}$ & $920.9^{4.1}$ \\
IAF$^{(8)}$ & $343.9^{1.4}$ & $498.8^{2.3}$ & $774.7^{2.9}$ & $885.7^{2.8}$ \\ \hline
HF$^{(1)}$\Tstrut & $335.5^{1.2}$ & $488.6^{2.0}$ & $795.9^{3.3}$ & $917.0^{2.4}$ \\
HF$^{(2)}$ & $340.6^{1.3}$ & $495.9^{1.8}$ & $784.5^{4.8}$ & $929.4^{3.7}$ \\
HF$^{(4)}$ & $343.3^{1.2}$ & $487.0^{2.7}$ & $799.7^{3.2}$ & $877.5^{4.7}$ \\
HF$^{(8)}$ & $343.3^{1.3}$ & $488.3^{2.4}$ & $794.6^{4.0}$ & $889.2^{4.7}$ \\ \hline
ME$^{(2)}$\Tstrut & $344.2^{1.5}$ & $491.7^{1.4}$ & $793.4^{3.8}$ & $880.3^{3.6}$ \\
ME$^{(3)}$ & $\color{dor} 350.3^{1.8}$ & $491.2^{2.1}$ & $807.5^{4.9}$ & $875.9^{4.6}$ \\
ME$^{(4)}$ & $337.7^{1.1}$ & $491.3^{1.8}$ & $732.0^{3.1}$ & $939.8^{8.6}$ \\
ME$^{(5)}$ & $343.0^{1.4}$ & $478.0^{2.8}$ & $805.7^{3.8}$ & $861.9^{7.0}$ \\ \hline
RME$^{(2)}$\Tstrut & $\color{dor} 349.3^{1.5}$ & $\color{dor} 508.2^{1.2}$ & $\color{red} {\bf 821.0}^{3.1}$ & $\color{dor} 941.5^{1.7}$ \\
RME$^{(3)}$ & $\color{dor} 349.9^{1.6}$ & $\color{dor} 507.5^{1.1}$ & $\color{dor} 820.4^{0.9}$ & $\color{red} {\bf 944.6}^{5.1}$ \\
RME$^{(4)}$ & $\color{dor} 350.7^{1.7}$ & $\color{dor} 509.0^{1.2}$ & $\color{dor} 819.9^{0.9}$ & $\color{dor} 944.4^{1.7}$ \\
RME$^{(5)}$ & $\color{red} {\bf 351.1}^{1.7}$ & $\color{red} {\bf 509.1}^{1.4}$ & $\color{dor} 819.9^{0.9}$ & $\color{dor} 944.0^{1.6}$ \\
\hline\hline
BVI-ER1$^{(2)}$\Tstrut & $\color{or} 349.2^{1.9}$ & $\color{lor} 507.9^{2.2}$ & $\color{lor} 817.1^{3.3}$ & $937.9^{5.1}$ \\
BVI-ER1$^{(3)}$ & $\color{or} 350.0^{1.9}$ & $\color{or} 507.8^{2.2}$ & $\color{lor} 816.6^{3.4}$ & $936.2^{5.1}$ \\
BVI-ER1$^{(4)}$ & $\color{dor} 350.7^{1.5}$ & $\color{dor} 507.8^{2.3}$ & $\color{lor} 816.8^{3.4}$ & $935.6^{3.8}$ \\
BVI-ER1$^{(5)}$ & $\color{dor} 351.1^{1.5}$ & $\color{dor} 508.2^{2.3}$ & $\color{lor} 816.4^{3.3}$ & $935.7^{3.8}$ \\ \hline
BVI-ER2$^{(2)}$\Tstrut & $\color{or} 349.3^{1.9}$ & $\color{lor} 507.8^{2.2}$ & $\color{lor} 817.1^{3.4}$ & $937.6^{5.1}$ \\
BVI-ER2$^{(3)}$ & $\color{or} 349.8^{1.9}$ & $\color{or} 507.8^{2.2}$ & $\color{lor} 816.6^{3.4}$ & $\color{lor} 936.1^{5.1}$ \\
BVI-ER2$^{(4)}$ & $\color{dor} 350.7^{1.5}$ & $\color{dor} 507.8^{2.2}$ & $\color{lor} 816.9^{3.4}$ & $935.6^{3.8}$ \\
BVI-ER2$^{(5)}$ & $\color{dor} 351.0^{1.5}$ & $\color{dor} 508.1^{2.2}$ & $\color{lor} 816.4^{3.4}$ & $935.7^{3.8}$ \\ % \hline
\bottomrule
\end{tabular}
%\end{sc}
\end{small}
\end{table}
%%%%

%%%%
\begin{table}%[H]
\centering
\caption{(CIFAR10) Test log-likelihood scores (unit in nat). The same interpretation as \autoref{tab:mnist}. 
}
\label{tab:cifar10}
%\vskip 0.05in
\begin{small}
%\begin{sc}
\centering
\begin{tabular}{lllll}
\toprule
$\textrm{dim}({\bf z})$ & \multicolumn{1}{c}{$10$} %$\textrm{dim}({\bf z})=10$ 
 & \multicolumn{1}{c}{$20$} & \multicolumn{1}{c}{$50$} & \multicolumn{1}{c}{$100$} \\
 %$\textrm{dim}({\bf z})=50$ & $\textrm{dim}({\bf z})=100$ \\
%\midrule
\hline\hline
VAE\Tstrut & $1645.7^{4.9}$ & $2089.7^{5.8}$ & $2769.9^{7.1}$ & $3381.0^{14.7}$ \\ \hline
SA$^{(1)}$\Tstrut & $1645.0^{5.6}$ & $2086.0^{6.2}$ & $2765.0^{7.1}$ & $3378.7^{10.4}$ \\
SA$^{(2)}$ & $1648.6^{4.8}$ & $2088.2^{6.6}$ & $2764.1^{7.7}$ & $3377.8^{9.8}$ \\
SA$^{(4)}$ & $1648.5^{5.2}$ & $2083.9^{8.4}$ & $2766.7^{6.6}$ & $3380.2^{7.9}$ \\
SA$^{(8)}$ & $1642.1^{5.4}$ & $2086.0^{6.1}$ & $2766.6^{7.5}$ & $3376.6^{10.6}$ \\ \hline
IAF$^{(1)}$\Tstrut & $1646.0^{4.9}$ & $2081.1^{5.4}$ & $2762.6^{7.2}$ & $3383.7^{7.1}$ \\
IAF$^{(2)}$ & $1642.0^{4.9}$ & $2084.6^{5.6}$ & $2763.0^{4.3}$ & $3373.3^{14.2}$ \\
IAF$^{(4)}$ & $1646.0^{5.1}$ & $2083.2^{6.1}$ & $2760.6^{7.0}$ & $3371.1^{8.1}$ \\
IAF$^{(8)}$ & $1643.6^{4.6}$ & $2087.1^{4.6}$ & $2761.8^{6.9}$ & $3364.0^{9.6}$ \\ \hline
HF$^{(1)}$\Tstrut & $1644.5^{4.4}$ & $2079.1^{5.5}$ & $2757.9^{4.4}$ & $3393.4^{4.7}$ \\
HF$^{(2)}$ & $1636.7^{4.9}$ & $2086.0^{5.9}$ & $2764.7^{4.4}$ & $3384.8^{4.7}$ \\
HF$^{(4)}$ & $1642.1^{4.9}$ & $2082.3^{7.3}$ & $2763.4^{4.4}$ & $3385.5^{4.4}$ \\
HF$^{(8)}$ & $1639.9^{5.4}$ & $2084.7^{6.1}$ & $2765.5^{7.2}$ & $3382.5^{4.3}$ \\ \hline
ME$^{(2)}$\Tstrut & $1643.6^{5.1}$ & $2086.6^{6.8}$ & $2767.9^{9.4}$ & $3378.5^{9.1}$ \\
ME$^{(3)}$ & $1638.6^{5.8}$ & $2079.8^{5.9}$ & $2770.2^{7.8}$ & $3388.1^{7.7}$ \\
ME$^{(4)}$ & $1641.8^{5.4}$ & $2084.7^{6.9}$ & $2763.5^{9.3}$ & $3384.6^{10.3}$ \\
ME$^{(5)}$ & $1641.7^{5.6}$ & $2080.2^{5.9}$ & $2766.1^{6.3}$ & $3351.3^{11.0}$ \\ \hline
RME$^{(2)}$\Tstrut & $\color{or} 1652.3^{5.0}$ & $\color{lor} 2095.7^{5.8}$ & $\color{dor} 2779.6^{6.6}$ & $\color{dor} 3403.0^{6.9}$ \\
RME$^{(3)}$ & $\color{dor} 1654.2^{4.9}$ & $\color{red} {\bf 2099.1}^{7.2}$ & $\color{red} {\bf 2783.0}^{6.1}$ & $\color{dor} 3404.2^{6.8}$ \\
RME$^{(4)}$ & $\color{red} {\bf 1655.0}^{6.4}$ & $\color{or} 2096.6^{5.9}$ & $\color{dor} 2781.1^{6.6}$ & $\color{dor} 3403.2^{6.1}$ \\
RME$^{(5)}$ & $\color{dor} 1654.5^{4.6}$ & $\color{dor} 2098.4^{5.8}$ & $\color{dor} 2782.9^{6.4}$ & $\color{red} {\bf 3404.6}^{5.7}$ \\
\hline\hline
BVI-ER1$^{(2)}$\Tstrut & $\color{lor} 1648.6^{5.1}$ & $\color{lor} 2094.4^{5.7}$ & $\color{lor} 2775.9^{6.4}$ & $3393.1^{6.8}$ \\
BVI-ER1$^{(3)}$ & $\color{lor} 1648.9^{5.0}$ & $\color{lor} 2094.7^{5.9}$ & $\color{lor} 2776.2^{6.6}$ & $3393.8^{6.5}$ \\
BVI-ER1$^{(4)}$ & $\color{lor} 1649.0^{5.1}$ & $\color{lor} 2095.0^{5.8}$ & $\color{lor} 2776.5^{6.3}$ & $3394.2^{6.6}$ \\
BVI-ER1$^{(5)}$ & $\color{lor} 1649.1^{5.2}$ & $\color{lor} 2095.1^{5.8}$ & $\color{lor} 2776.8^{6.5}$ & $3394.2^{7.7}$ \\ \hline
BVI-ER2$^{(2)}$\Tstrut & $\color{lor} 1648.6^{5.1}$ & $\color{lor} 2094.4^{5.7}$ & $\color{lor} 2775.8^{6.8}$ & $3393.1^{6.6}$ \\
BVI-ER2$^{(3)}$ & $\color{lor} 1648.9^{5.0}$ & $\color{lor} 2094.7^{5.7}$ & $\color{lor} 2776.2^{6.6}$ & $3393.8^{6.5}$ \\
BVI-ER2$^{(4)}$ & $\color{lor} 1649.0^{5.1}$ & $\color{lor} 2095.0^{5.8}$ & $\color{lor} 2776.5^{6.3}$ & $3394.2^{6.2}$ \\
BVI-ER2$^{(5)}$ & $\color{lor} 1649.1^{5.1}$ & $\color{lor} 2095.1^{5.8}$ & $\color{lor} 2776.8^{6.5}$ & $3394.1^{6.1}$ \\ % \hline
\bottomrule
\end{tabular}
%\end{sc}
\end{small}
\end{table}
%%%%

%%%%
\begin{table}%[H]
\centering
\caption{(SVHN) Test log-likelihood scores (unit in nat). The same interpretation as \autoref{tab:mnist}. 
}
\label{tab:svhn}
%\vskip 0.05in
\begin{small}
%\begin{sc}
\centering
\begin{tabular}{lllll}
\toprule
$\textrm{dim}({\bf z})$ & \multicolumn{1}{c}{$10$} %$\textrm{dim}({\bf z})=10$ 
 & \multicolumn{1}{c}{$20$} & \multicolumn{1}{c}{$50$} & \multicolumn{1}{c}{$100$} \\
 %$\textrm{dim}({\bf z})=50$ & $\textrm{dim}({\bf z})=100$ \\
%\midrule
\hline\hline
VAE\Tstrut & $3360.2^{9.1}$ & $4054.5^{14.3}$ & $5363.7^{21.4}$ & $6703.0^{28.4}$ \\ \hline
SA$^{(1)}$\Tstrut & $3358.7^{8.9}$ & $4031.5^{19.0}$ & $5362.1^{35.7}$ & $6707.6^{24.8}$ \\
SA$^{(2)}$ & $3356.0^{8.8}$ & $4041.5^{15.5}$ & $5377.0^{23.2}$ & $6697.0^{35.5}$ \\
SA$^{(4)}$ & $3327.8^{8.2}$ & $4051.9^{22.2}$ & $5391.7^{20.4}$ & $6645.1^{19.8}$ \\
SA$^{(8)}$ & $3352.8^{11.5}$ & $4041.6^{9.5}$ & $5370.8^{18.5}$ & $6674.5^{20.9}$ \\ \hline
IAF$^{(1)}$\Tstrut & $\color{lor} 3377.1^{8.4}$ & $4050.0^{9.4}$ & $5368.3^{11.5}$ & $6650.3^{15.7}$ \\
IAF$^{(2)}$ & $3362.3^{8.9}$ & $4054.6^{10.5}$ & $5360.0^{10.0}$ & $6671.5^{16.8}$ \\
IAF$^{(4)}$ & $3346.1^{8.7}$ & $4048.6^{8.7}$ & $5338.1^{10.2}$ & $6630.0^{17.2}$ \\
IAF$^{(8)}$ & $\color{lor} 3372.6^{8.3}$ & $4042.0^{9.6}$ & $5341.8^{10.1}$ & $6602.0^{10.8}$ \\ \hline
HF$^{(1)}$\Tstrut & $\color{or} 3381.4^{8.9}$ & $4028.8^{9.7}$ & $5372.0^{10.1}$ & $6678.8^{8.8}$ \\
HF$^{(2)}$ & $3342.4^{8.3}$ & $4030.7^{9.9}$ & $5376.6^{10.2}$ & $6672.0^{9.6}$ \\
HF$^{(4)}$ & $\color{lor} 3370.0^{8.2}$ & $4038.4^{9.7}$ & $5371.8^{9.8}$ & $6655.2^{9.5}$ \\
HF$^{(8)}$ & $3343.8^{8.2}$ & $4035.9^{8.9}$ & $5351.1^{11.1}$ & $6642.4^{16.5}$ \\ \hline
ME$^{(2)}$\Tstrut & $3352.3^{9.9}$ & $4037.2^{11.0}$ & $5343.2^{13.1}$ & $6670.2^{46.5}$ \\
ME$^{(3)}$ & $3335.2^{10.9}$ & $4053.8^{16.1}$ & $5367.7^{15.8}$ & $6605.6^{9.4}$ \\
ME$^{(4)}$ & $3358.2^{14.9}$ & $4061.3^{12.0}$ & $5191.9^{18.5}$ & $6605.7^{9.2}$ \\
ME$^{(5)}$ & $3360.6^{7.8}$ & $4057.5^{12.2}$ & $5209.2^{12.8}$ & $6604.0^{16.6}$ \\ \hline
RME$^{(2)}$\Tstrut & $\color{dor} 3390.0^{8.1}$ & $\color{dor} 4085.3^{9.7}$ & $\color{or} 5403.2^{10.2}$ & $\color{red} {\bf 6784.7}^{25.0}$ \\
RME$^{(3)}$ & $\color{red} {\bf 3392.0}^{12.6}$ & $\color{dor} 4085.9^{9.8}$ & $\color{dor} 5405.1^{10.4}$ & $\color{lor} 6782.7^{9.3}$ \\
RME$^{(4)}$ & $\color{dor} 3388.6^{8.3}$ & $\color{dor} 4080.7^{9.9}$ & $\color{or} 5403.8^{10.2}$ & $\color{lor} 6780.2^{9.4}$ \\
RME$^{(5)}$ & $\color{dor} 3391.9^{8.2}$ & $\color{red} {\bf 4086.9}^{10.9}$ & $\color{red} {\bf 5405.5}^{8.5}$ & $\color{lor} 6781.8^{10.0}$ \\
\hline\hline
BVI-ER1$^{(2)}$\Tstrut & $\color{dor} 3379.9^{8.2}$ & $\color{lor} 4077.3^{10.3}$ & $5388.2^{10.2}$ & $6753.5^{10.0}$ \\
BVI-ER1$^{(3)}$ & $\color{dor} 3380.9^{8.1}$ & $\color{lor} 4076.6^{10.3}$ & $5384.2^{10.5}$ & $6750.3^{10.6}$ \\
BVI-ER1$^{(4)}$ & $\color{dor} 3384.4^{8.1}$ & $\color{lor} 4073.1^{10.2}$ & $5371.1^{10.4}$ & $6748.9^{11.3}$ \\
BVI-ER1$^{(5)}$ & $\color{dor} 3382.2^{8.4}$ & $\color{lor} 4071.2^{10.2}$ & $5378.1^{10.1}$ & $6733.6^{15.3}$ \\ \hline
BVI-ER2$^{(2)}$\Tstrut & $\color{dor} 3379.8^{8.1}$ & $\color{lor} 4077.3^{9.8}$ & $5388.3^{10.1}$ & $6753.2^{10.1}$ \\
BVI-ER2$^{(3)}$ & $\color{dor} 3380.9^{8.4}$ & $\color{lor} 4076.7^{9.6}$ & $5383.9^{10.2}$ & $6749.7^{10.7}$ \\
BVI-ER2$^{(4)}$ & $\color{dor} 3384.3^{8.2}$ & $\color{lor} 4073.2^{9.2}$ & $5371.3^{10.4}$ & $6749.1^{11.1}$ \\
BVI-ER2$^{(5)}$ & $\color{dor} 3382.1^{8.4}$ & $\color{lor} 4071.2^{10.4}$ & $5377.7^{10.2}$ & $6733.8^{15.0}$ \\ % \hline
\bottomrule
\end{tabular}
%\end{sc}
\end{small}
\end{table}
%%%%

%%%%
\begin{table}%[t]
\centering
\caption{(CelebA) Test log-likelihood scores (unit in nat). The same interpretation as \autoref{tab:mnist}. 
}
\label{tab:celeba}
%\vskip 0.05in
\begin{small}
%\begin{sc}
\centering
\begin{tabular}{lllll}
\toprule
$\textrm{dim}({\bf z})$ & \multicolumn{1}{c}{$10$} %$\textrm{dim}({\bf z})=10$ 
 & \multicolumn{1}{c}{$20$} & \multicolumn{1}{c}{$50$} & \multicolumn{1}{c}{$100$} \\
 %$\textrm{dim}({\bf z})=50$ & $\textrm{dim}({\bf z})=100$ \\
%\midrule
\hline\hline
VAE\Tstrut & $9767.7^{36.0}$ & $12116.4^{25.3}$ & $15251.9^{39.7}$ & $17395.5^{32.4}$ \\ \hline
SA$^{(1)}$\Tstrut & $9735.2^{21.4}$ & $12091.1^{21.6}$ & $15285.8^{29.4}$ & $17432.4^{30.4}$ \\
SA$^{(2)}$ & $9754.2^{20.4}$ & $12087.1^{21.5}$ & $15252.7^{29.0}$ & $17434.0^{29.8}$ \\
SA$^{(4)}$ & $9769.1^{20.6}$ & $12116.3^{20.5}$ & $15187.3^{27.9}$ & $17360.5^{28.9}$ \\
SA$^{(8)}$ & $9744.8^{19.4}$ & $12100.6^{22.8}$ & $15096.5^{27.2}$ & $17409.7^{28.0}$ \\ \hline
IAF$^{(1)}$\Tstrut & $9750.3^{27.4}$ & $12098.0^{20.6}$ & $15271.2^{28.6}$ & $17446.4^{30.3}$ \\
IAF$^{(2)}$ & $9794.4^{23.3}$ & $12104.5^{21.8}$ & $15262.2^{27.8}$ & $17449.5^{31.8}$ \\
IAF$^{(4)}$ & $9764.7^{29.5}$ & $12094.6^{22.6}$ & $15261.0^{28.1}$ & $17416.8^{29.8}$ \\
IAF$^{(8)}$ & $9764.0^{21.6}$ & $12109.3^{22.0}$ & $15241.5^{27.9}$ & $17452.5^{39.5}$ \\ \hline
HF$^{(1)}$\Tstrut & $9748.3^{29.5}$ & $12077.2^{31.4}$ & $15240.5^{27.6}$ & $17461.6^{29.9}$ \\
HF$^{(2)}$ & $9765.8^{25.6}$ & $12093.0^{25.6}$ & $15258.2^{30.3}$ & $17479.8^{30.0}$ \\
HF$^{(4)}$ & $9754.3^{23.8}$ & $12082.0^{27.0}$ & $15266.5^{29.5}$ & $17532.7^{30.6}$ \\
HF$^{(8)}$ & $9737.5^{24.5}$ & $12087.3^{25.5}$ & $15248.7^{29.7}$ & $17663.4^{28.7}$ \\ \hline
ME$^{(2)}$\Tstrut & $\color{dor} 9825.3^{20.7}$ & $12072.7^{23.3}$ & $15290.5^{29.3}$ & $17419.3^{28.7}$ \\
ME$^{(3)}$ & $9797.6^{22.3}$ & $12100.3^{21.7}$ & $15294.6^{28.3}$ & $17395.3^{28.9}$ \\
ME$^{(4)}$ & $\color{dor} 9834.9^{25.4}$ & $12092.2^{22.6}$ & $15270.7^{20.6}$ & $17458.5^{36.8}$ \\
ME$^{(5)}$ & $9717.0^{23.2}$ & $12095.3^{25.1}$ & $15268.8^{27.5}$ & $17406.8^{31.8}$ \\ \hline
RME$^{(2)}$\Tstrut & $\color{dor} 9837.9^{24.6}$ & $\color{dor} 12193.1^{23.5}$ & $\color{dor} 15363.0^{31.7}$ & $\color{dor} 17873.5^{32.8}$ \\
RME$^{(3)}$ & $\color{dor} 9838.5^{25.0}$ & $\color{dor} 12192.3^{23.5}$ & $\color{dor} 15365.6^{31.4}$ & $\color{dor} 17874.4^{31.2}$ \\
RME$^{(4)}$ & $\color{red} {\bf 9849.5}^{12.1}$ & $\color{dor} 12192.6^{23.4}$ & $\color{dor} 15364.3^{31.5}$ & $\color{red} {\bf 17875.1}^{14.2}$ \\
RME$^{(5)}$ & $\color{dor} 9843.5^{25.0}$ & $\color{red} {\bf 12194.2}^{11.5}$ & $\color{red} {\bf 15366.2}^{12.7}$ & $\color{dor} 17874.3^{32.5}$ \\
\hline\hline
BVI-ER1$^{(2)}$\Tstrut & $9801.6^{26.1}$ & $12133.5^{25.1}$ & $15206.4^{28.2}$ & $17716.9^{70.3}$ \\
BVI-ER1$^{(3)}$ & $9805.6^{25.7}$ & $12146.5^{22.4}$ & $15249.5^{28.1}$ & $17558.6^{120.1}$ \\
BVI-ER1$^{(4)}$ & $9805.2^{29.3}$ & $12127.7^{22.3}$ & $15085.8^{28.4}$ & $17256.1^{283.9}$ \\
BVI-ER1$^{(5)}$ & $9810.1^{30.7}$ & $12092.3^{22.3}$ & $15052.5^{28.0}$ & $17069.9^{391.8}$ \\ \hline
BVI-ER2$^{(2)}$\Tstrut & $9801.5^{25.3}$ & $12133.6^{28.7}$ & $15207.3^{52.4}$ & $17716.6^{92.1}$ \\
BVI-ER2$^{(3)}$ & $9805.7^{24.9}$ & $12146.6^{25.5}$ & $15249.6^{54.6}$ & $17560.7^{109.2}$ \\
BVI-ER2$^{(4)}$ & $9805.1^{26.3}$ & $12128.7^{34.0}$ & $15084.9^{42.5}$ & $17260.6^{228.6}$ \\
BVI-ER2$^{(5)}$ & $9810.4^{27.8}$ & $12087.5^{48.9}$ & $15051.7^{43.5}$ & $17077.1^{387.6}$ \\ % \hline
\bottomrule
\end{tabular}
%\end{sc}
\end{small}
\end{table}
%%%%

\clearpage

%\section*{References}

%\small
{
\small
\bibliography{main}
\bibliographystyle{ieee_fullname}
}

% References follow the acknowledgments. Use unnumbered first-level heading for
% the references. Any choice of citation style is acceptable as long as you are
% consistent. It is permissible to reduce the font size to \verb+small+ (9 point)
% when listing the references.
% {\bf Note that the Reference section does not count towards the eight pages of content that are allowed.}
% \medskip

% \small

% [1] Alexander, J.A.\ \& Mozer, M.C.\ (1995) Template-based algorithms for
% connectionist rule extraction. In G.\ Tesauro, D.S.\ Touretzky and T.K.\ Leen
% (eds.), {\it Advances in Neural Information Processing Systems 7},
% pp.\ 609--616. Cambridge, MA: MIT Press.

% [2] Bower, J.M.\ \& Beeman, D.\ (1995) {\it The Book of GENESIS: Exploring
%   Realistic Neural Models with the GEneral NEural SImulation System.}  New York:
% TELOS/Springer--Verlag.

% [3] Hasselmo, M.E., Schnell, E.\ \& Barkai, E.\ (1995) Dynamics of learning and
% recall at excitatory recurrent synapses and cholinergic modulation in rat
% hippocampal region CA3. {\it Journal of Neuroscience} {\bf 15}(7):5249-5262.

\end{document}